\documentclass[conference]{IEEEtran}
\usepackage{times}
\usepackage{booktabs}
\usepackage{multirow}
\usepackage{graphicx}
\usepackage{amsmath}
\usepackage{amssymb}
\usepackage{algorithm}
\usepackage{algpseudocode}
\usepackage{gensymb}
\usepackage{xcolor}
\usepackage{tablefootnote}
\usepackage{makecell}
\usepackage{twemojis}
\let\labelindent\relax
\usepackage[inline]{enumitem}

\usepackage[numbers]{natbib}
\usepackage{multicol}
\usepackage[bookmarks=true]{hyperref}

\usepackage{xcolor}
\usepackage[export]{adjustbox}
\definecolor{darkgreen}{rgb}{0.0, 0.2, 0.13}

\newcommand{\methodname}{\textsc{SafeMimic}}

\newcommand{\arpit}[1]{\textcolor{black}{#1}}

\begin{document}

\title{SafeMimic: Towards Safe and Autonomous Human-to-Robot Imitation for Mobile Manipulation}

\author{Arpit Bahety, Arnav Balaji, Ben Abbatematteo, Roberto Martín-Martín \\
The University of Texas at Austin}



%

\maketitle
\IEEEpeerreviewmaketitle

\begin{abstract}
For robots to become efficient helpers in the home, they must learn to perform new mobile manipulation tasks simply by watching humans perform them. 
Learning from a single video demonstration from a human is challenging as the robot needs to first extract from the demo \textit{what} needs to be done and \textit{how}, translate the strategy from a third to a first-person perspective, and then adapt it to be successful with its own morphology.
Furthermore, to mitigate the dependency on costly human monitoring, this learning process should be performed in a safe and autonomous manner. 
We present \methodname{}, a framework to learn new mobile manipulation skills safely and autonomously from a single third-person human video.
Given an initial human video demonstration of a multi-step mobile manipulation task, \methodname{} first parses the video into segments, inferring both the semantic changes caused and the motions the human executed to achieve them and translating them to an egocentric reference. 
Then, it adapts the behavior to the robot's own morphology by sampling candidate actions around the human ones, and verifying them for safety before execution in a receding horizon fashion using an ensemble of safety Q-functions trained in simulation. 
When safe forward progression is not possible, \methodname{} \textit{backtracks} to previous states and attempts a different sequence of actions, adapting both the trajectory and the grasping modes when required for its morphology. 
As a result, \methodname{} yields a strategy that 
succeeds in the demonstrated behavior and learns task-specific actions that reduce exploration in future attempts.
Our experiments show that our method allows robots to safely and efficiently learn multi-step mobile manipulation behaviors from a single human demonstration, from different users, and in different environments, with improvements over state-of-the-art baselines across seven tasks. For more information and video results, \url{https://robin-lab.cs.utexas.edu/SafeMimic/}
\end{abstract}
\section{Introduction}

For decades, we have dreamed to teach robots a new task in the same way we would teach it to another human: by demonstrating it in from of them.
This would bypass the need for costly teleoperated data collection~\cite{o2024open, khazatsky2024droid, ebert2021bridge}, which is significantly complex and time-consuming for multi-step tasks and those combining navigation and manipulation into mobile manipulation.
Recent advances in human motion perception and parsing~\cite{ye2023slahmr, pavlakos2024reconstructing, Shan20, rong2021frankmocap} have brought us closer to this dream:
they have enabled new approaches that extract information from human videos and use it to ``seed'' an exploratory process in which the robot adapts the strategy to its own embodiment~\cite{bahl2022human, kannan2023deft, shaw2023videodex, bahety_screwmimic_2024}.
However, these techniques are restricted to \textbf{short horizon} skills and require tedious \textbf{human supervision}, tasked with ensuring that the robot exploration is \textbf{safe}, \textbf{resetting} the task constantly and \textbf{detecting success}.
What is necessary to enable robots to learn multi-step mobile manipulation tasks from a single human demonstration \textbf{safely} and \textbf{autonomously}?



\begin{figure}[t]
    \centering
    \includegraphics[width=1\columnwidth, trim={0 1cm 0 0},clip]{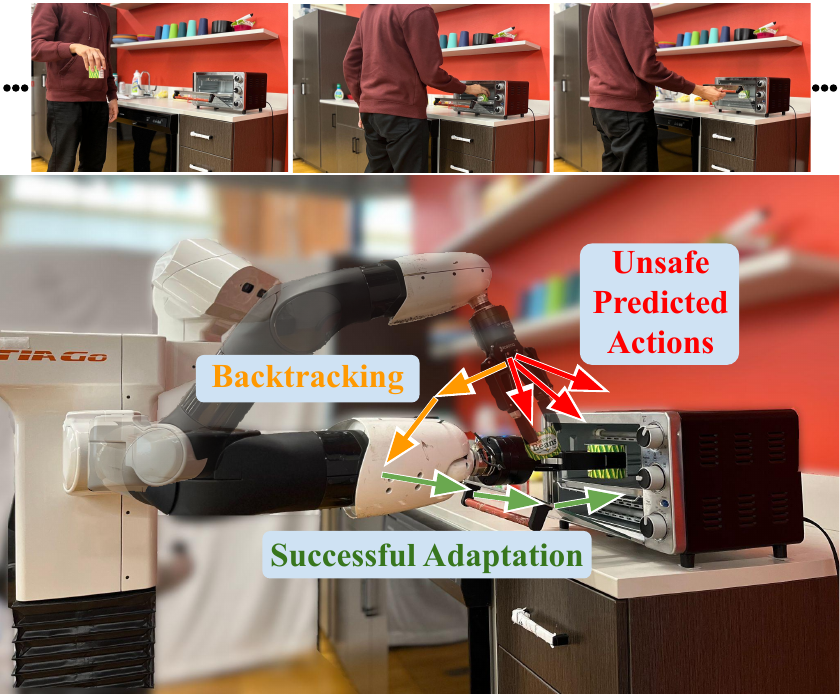}
    \caption{Robot imitating a single video of a human-demonstrated mobile manipulation task safely and autonomously with \methodname{}. From a video of a multi-step mobile manipulation task (\textit{top}), \methodname{} extracts an initial motion strategy and adapts it to its own embodiment by exploring new actions in a safe manner. It combines an ensemble of safety Q-functions that predict future unsafe motions (\textit{\textcolor{red}{red arrows}}, motion will collide) and a backtracking mechanism (\textit{\textcolor{orange}{orange arrows}}) that enables autonomous exploration until a successful action sequence is found (\textit{\textcolor{teal}{green arrows}}).}
    \label{fig:pullfig} 
    \vspace*{-2em}
\end{figure}


Learning multi-step tasks from a human video in a safe and self-supervised manner presents multiple technical challenges.
First, it requires for the robot to understand both the high-level semantics of the task and the associated low-level motion. 
This implies extracting from the video both the human's motion as well as the semantic changes they caused in the environment.
This information needs to be translated from a third to a first person view so that the robot can execute the motion and monitor the semantic changes with its onboard sensors.
When executed, the initial translated motion may fail due to morphological differences or noise in the video parsing; while finding small adaptations have been shown feasible through trial-and-error learning in the real world~\cite{bahl2022human, kannan2023deft, shaw2023videodex, bahety_screwmimic_2024}, it is still unclear how to explore for longer multi-step tasks that require a larger adaptation, e.g., for different grasping strategies.
Finally, such a real-world exploration quickly becomes unsafe due to potential damage to the robot or the environment, requiring another agent (human) to monitor and reset the scene for further exploration. 
Safe and autonomous exploration calls for a new method that allows the robot to predict when something may go wrong before it happens and backtrack to previous states to keep trying new strategies.

In this work, we introduce \methodname{}, a framework for safely and autonomously learning  mobile manipulation behaviors from a single third-person human video.
\methodname{} overcomes all aforementioned challenges: first, it parses the third-person demonstration combining a human motion tracker and a vision-language model (VLM), obtaining an initial task plan composed of sequences of distinct human motion and easy-to-detect semantic changes executable from a first-person perspective. 
Then, \methodname{} adapts the initial plan by refining each segment using a safe exploration procedure in the real world. 
At the core of our safe exploration sits an ensemble of \textbf{safety Q-functions} pretrained in simulation that enable \methodname{} to attempt risk-free actions both for continuous motion as well as discrete grasps, overcoming large differences in morphology and manipulation capabilities. 
\arpit{General safety Q-functions have the potential to generalize across tasks as they capture generic risks common to manipulation --such as collisions, excessive forces, or grasp losses-- and, as we observe empirically, they require less precise sim-to-real alignment than direct policy transfer.}
When \methodname{} detects no safe actions to progress in the exploration, it backtracks to a previous state and tries a different strategy, including changing the grasping mode if necessary. 
Finally, when a safe and successful adaptation is found, \methodname{} associates it to the geometry of the task so that exploration is reduced in future attempts.

In summary, \methodname{} introduces several novel contributions: 
\begin{itemize}[leftmargin=*]
    \item A comprehensive framework to parse a single multi-step mobile manipulation video demonstration from a human and adapt it to the robot's capabilities in a safe and self-supervised manner.
    \item A mechanism to parse human videos into sequences of motions with unique and easy-to-perceive semantic effects combining human pose tracking and VLM detections. 
    \item A safe exploration in the real world with a predictive control strategy informed by an ensemble of safety Q-functions pretrained in simulation. 
    \item A self-supervised mechanism to detect success  and to backtrack to previous states to try new strategies, including different grasp modes.
    \item A mechanism to learn from previous experiences to reduce the amount of exploration necessary in future attempts.
\end{itemize}

We demonstrate the performance of \methodname{} in seven challenging, multi-step mobile manipulation tasks in different environments with different human teachers, and observe experimentally that our framework enables the robot to successfully acquire the desired behaviors safely and more efficiently than direct sim-to-real imitation learning approaches, previous human-to-robot methods, and variants without safety Q-functions. 

\section{Related Work} 
\label{sec:rew}
\methodname{} is a novel framework for safely and autonomously learning multi-step mobile manipulation tasks from human demonstrations. In this section, we contrast \methodname{} to prior efforts on learning from human video as well as safe imitation and reinforcement learning. 



\textbf{Learning from Human Video} directly has received increasing attention as strategy to learn manipulation skills. 
Some works have explored leveraging large collections of human activity data~\cite{grauman2022ego4d, sivakumar2022robotic, xiong2021learning} in order to learn cost functions from video and language data~\cite{chen2021learning, shao2021concept2robot, ma2023liv, ma2022vip}. Other works have explored human video modeling as a pretraining objective~\cite{nair2022r3m, radosavovic2023real}. 
Most closely related to \methodname{}, several works imitate human actions directly by tracking the human pose and extracting actions using pose tracking before finetuning with gradient-free RL~\cite{bahl2022human, bharadhwaj2023zeroshot, kannan2023deft, shaw2023videodex, mendonca2023structured, bahety_screwmimic_2024}. 
These works demonstrate impressive behaviors, yet typically refine the initial policy obtained from the human in a naive exploratory fashion. This has the potential to damage the robot or environment, as actions are not checked for collision or other potentially dangerous failures. Other works rely on morphological similarity between humans and humanoid robots to retarget motions directly~\cite{li2024okami, cheng2024expressive, choi2020nonparametric, he2024learning}. In contrast, we focus on imitating and refining the demonstration safely when the robot and human embodiments do not necessarily match.  Moreover, most existing works on learning from human video focus on short-horizon tasks (e.g., opening a drawer) rather than multi-step mobile manipulation behaviors. 

\textbf{Autonomous Real-World Learning}
prior methods have examined learning to automatically reset the environment in order to enable learning without human supervision~\cite{zhu2020ingredients, mendonca2024continuously, sharma2023self, nguyen2023provable, walke2023don, tang2024deep}. These works address the requirement of autonomy, but generally sidestep the question of safety --- a critical challenge when learning mobile manipulation in the real world. Further, these methods require extensive trial-and-error learning not  suitable for efficiently learning from a single human video demonstration.  \methodname{} instead combines both autonomous and safe exploration when learning from human video and employs a simple but effective backtracking strategy when failures are predicted for all sampled actions in a given state. 

\textbf{Safe Imitation and Reinforcement Learning} has received significant study in order to reduce risks in the real world. \textit{Design-time} approaches~\cite{ross2011reduction, laskey2017dart, ciftci2024safe, liu2024model, xuuncertainty, gokmen2023asking} operate during data collection or model training in order to ensure robustness to perturbation during execution. 
When learning from human video, design-time approaches are infeasible, as they are typically targeted for teleoperation when disturbances can be injected. 
\textit{Deployment-time} approaches~\cite{reichlin2022back, wabersich2023data, hsu2023safety, yang2024enhancing,wong2022error} filter actions and defer to a backup or recovery policy when constraints are violated. 
Deployment-time methods usually assume access to constraints in closed form, an unrealistic requirement in novel environments and tasks. \methodname{} employs learned safety Q-functions that are pretrained from simulation data, allowing it to learn from human video demonstrations safely.  


\begin{figure*}[t!]
    \centering
    \includegraphics[width=\textwidth]{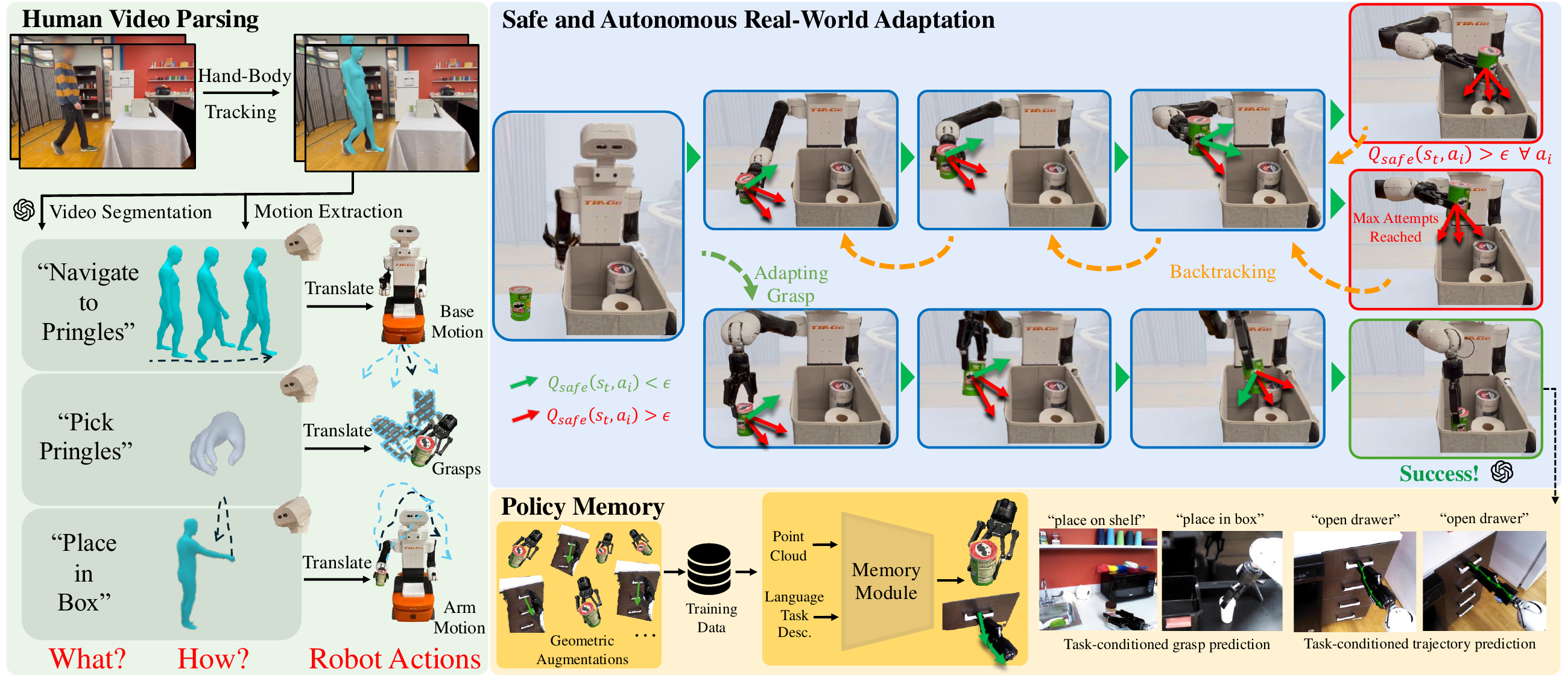}
    \caption{\textbf{Overview of \methodname{}.} From an RGB-D video of a human performing a multi-step mobile manipulation task acquired by the robot, \methodname{} uses a combination of human pose tracking models~\cite{ye2023slahmr, pavlakos2024reconstructing} and VLM prompting to perform coarse-to-fine segmentation obtaining semantic changes --``what?''-- and human action trajectories --``how?''--, and translating them to the robot's point of view (\textit{left}, Sec.~\ref{ss:ta}). 
    \methodname{} then refines and adapts each task segment safely by sampling and verifying actions before executing them thanks to an ensemble of safety Q-functions pretrained in simulation, $Q_\textrm{safe}$ (\textit{top right}, Sec.~\ref{ss:tb}). 
    If forward progress is not possible, \methodname{} autonomously backtracks and tries different actions (\textit{\textcolor{orange}{orange arrows}}). 
    When required to overcome large differences in morphology, \methodname{} explores alternative grasp modes (\textit{second row of samples}), adapting the grasp to enable successful execution. 
    Successful attempts are detected by a VLM that verifies when the parsed semantic change of the segment is achieved. Successes are stored and used to train a policy memory module with geometric augmentations  (\textit{bottom right}, Sec.~\ref{ss:tc}) that predicts actions (grasps or action trajectories) in subsequent attempts, given a pointcloud and language task description, in order to reduce the need for exploration.}
    \label{fig:method_diagram} 
    \vspace*{-1.5em}
\end{figure*}

Safe RL methods~\cite{alshiekh2018safe, srinivasan2020learning, thananjeyan2021recovery, yang2022safe} provide a framework for safe policy learning, enabling the discover of complex behaviors from scratch while avoiding failures in a target task. 
These approaches typically jointly optimize task and safety Q-functions during a pre-training phase, and do not address the case of imitation learning or learning from human video. In contrast, we employ task-agnostic data collection in simulation, and learn directly from human teachers in the real world. 
Constrained RL methods~\cite{liu2022robot, bouvier_policed_2024, Dalal2018SafeEI, yu2022reachability} similarly allow for policy learning while obeying constraints, though typically require closed-form constraints available at runtime. In the real-world, such constraints are difficult to acquire in novel environments, much less in the presence of human teachers. 

\textbf{Failure Prediction} ahead of execution has received considerable study in robotics, often with the goal of providing safety certificates on policies for deployment~\cite{bharadhwaj2022auditing, twala_robot_2009, alvanpour_robot_2020, diehl_causal-based_2023}. Many works approach failure prediction through the lens of reachability analysis~\cite{akametalu2014reachability, mitchell2005time} or design control barrier functions~\cite{ames2016control}.
Recent work has sought to learn failure predictors with PAC guarantees~\cite{farid2022failure} or conformal prediction~\cite{luo2024sample}. 
However, these approaches assume access to a black box policy or dynamics model of the environment, both which are unknown in the case of learning a new task from human video in a novel environment. 
Similarly, motion planning methods~\cite{la2011motion, chitta2012moveit} enable collision-free motion generation for a given environment geometry but fail to capture other possible failure modes involving contact, such as force-torque limit violations or grasp loss. \methodname{} provides a unified framework for failure prediction when learning mobile manipulation behaviors from human videos.

\section{\methodname{} Framework} 
\label{sec:method}

Our goal is to enable a robot to safely and autonomously learn to adapt and imitate a multi-step mobile manipulation task demonstrated in a single third-person human video. 
Figure~\ref{fig:method_diagram} provides an overview of our framework, \methodname{}.
In this section, we describe each of its components. 

\subsection{Factorizing, Parsing and Translating Human Videos}
\label{ss:ta}
The first step to imitate a human video demonstration is to extract information from it: \textit{what} did the human do, and \textit{how} did they do it?
In \methodname{}, we identify the \textit{what} by detecting the semantic changes caused by the human in the environment, and the \textit{how} by tracking the motion of the human that caused those changes.
However, as we aim at imitating complex multi-step mobile manipulation tasks, extracting  multiple semantic changes and intricate human motion for the entire video and learning to adapt it to the robot all at once may be unfeasible.
\methodname{} factorizes the original video into distinct segments where only a single semantic change happens.
This naturally leads to segments with separate navigation and manipulation to achieve single semantic changes such as \texttt{navigate\_to}, \texttt{reach\_for\_and\_grasp}, \texttt{open}, \ldots (see full list in Appendix~\ref{sec:app_segmentation}) that can be adapted and optimized sequentially in a self-supervised manner by the robot.

To factorize and parse the multi-step video into single semantic change segments, \methodname{} combines a body and hand visual tracker~\cite{ye2023slahmr,pavlakos2024reconstructing}, a contact detector~\cite{Shan20}, and a VLM~\cite{achiam2023gpt}.
An initial coarse segmentation is obtained based on the tracked human motion by detecting if the human is navigating or manipulating by thresholding the amount of inter-frame body translation.
Then, \methodname{} annotates each coarse segment with its semantic change and possibly factories it further. 
Navigation segments are assumed to be at the lowest level of granularity; \methodname{} just needs to extract the semantic goal, the object/location that the navigation segment tries to reach, obtained using a VLM.
Manipulation segments are further factorized by \methodname{} as necessary by combining information from a VLM and a contact detector that separates manipulation phases without contact from the ones with contact. 
This leads to a natural factorization into segments that begin with a grasping action or with a change in the contact interface, e.g., when a wiping segment begins.
Each segment is then annotated with its semantic goal, again obtained with a VLM query.

The factorization and parsing process mentioned above leads to a sequence of segments with a individual semantic goals and motions of the body and/or arm of the human (Fig.~\ref{fig:method_diagram}, left).
While the semantic change is invariant to the point of view, the motion is viewed from a third-person perspective and needs to be translated into the robot's reference frame to be executed and monitored with the onboard sensors. Similarly, the human grasps may need to be translated to the robot morphology.
To that end, \methodname{} assumes that in navigation segments, the robot starts from a location \textit{close enough} to the human initial location (no calibration needed!) and translates the sequence of human navigation actions to relative changes in base pose between frames.
Similarly, for manipulation segments, \methodname{} first transforms the hand pose to be relative to the human body and then computes the relative motion of the hand between consecutive frames. 
To translate grasps, \methodname{} will detect grasp candidates available to the robot~\cite{ten2017grasp} and match the one that is closest to the grasp demonstrated by the human. 
However, other grasp candidates will be explored as a result of our safe and autonomous human-to-robot adaptation, as we explain in the next section.

\subsection{Safe and Autonomous Real-World Adaptation}
\label{ss:tb}

The factorizing, parsing, and translating process described above leads to an initial policy that the robot could directly execute.
However, this initial policy will fail due to differences in embodiment and tracking inaccuracies (see Exp.~\ref{sec:results}).
\methodname{} implements an exploration and adaptation procedure in the real world to find the actions (close to the ones demonstrated by the human) that will lead to the same sequence of semantic changes and thus to success on the multi-step mobile manipulation task.
To that end, \methodname{} explores actions for each of the segments in turn, until the semantic goal of the segment is achieved in a process summarized in pseudocode in Alg.~\ref{algo:algo}.
In contrast to previous human-to-robot approaches, \methodname{} will perform this real-world exploration \textit{safely and autonomously} thanks to a combination of safety Q-functions and backtracking capabilities (see Fig.~\ref{fig:method_diagram}, right).

\textit{Safe Exploration with Safety Q-functions:} We assume that, at each segment, the robot's objective is to achieve the segment's semantic goal while avoiding unsafe states. 
We adopt the standard MDP formalism and represent each segment by the tuple $\mathcal{M} = (\mathcal{S}, \mathcal{A}, R, T, \gamma)$, 
where $\mathcal{S}$ is the state space, $\mathcal{A}$ is the action space, $R$ is a sparse reward based on task success (e.g., semantic goal achievement), $T$ is the transition function, and $\gamma$ is a discount factor. 
Using the Safe RL framework proposed by~\citet{srinivasan2020learning}, we denote the set of unsafe states $\mathcal{S}_{\text{unsafe}} = { s \in \mathcal{S} \mid \mathcal{I}(s) = 1 }$, where $\mathcal{I}(s)$ is an indicator boolean function that triggers for any unsafe states. This function can then be  considered a composition of single failure mode indicators, $\mathcal{I}(s)=\max\limits_{i} \mathcal{I}_i(s)$, where $\{\mathcal{I}_i\}$ a set of functions for distinct failure modes, e.g., exerting too much force, colliding with the environment, reaching joint limits, \ldots (for the complete list of unsafe state, see Appendix~\ref{sec:app_details}). 
Given this function, the robot's objective is to find a policy that maps states to the actions that maximize the task reward while remaining safe, given formally by:
$$
\max _\pi \sum_{t=0}^T \mathbb{E}_{\left(s_t, a_t\right) \sim \rho_\pi}\left[R\left(s_t, a_t\right)\right] \ \text{s.t.} \  \mathbb{E}_{s_t \sim \rho_\pi}\left[\mathcal{I}\left(s_t\right)\right] = 0,$$
where $\rho_\pi$ denotes the state-action distribution visited by the policy $\pi$.

Thus, to satisfy this objective, the robot requires a predictive model of the actions that will result in failure. 
To achieve this, we define a \textit{Safety Q-function} that predicts the probability that an action $a_t$ taken in state $s_t$ will result in a failure: 
$$Q_{\mathrm{safe}}\left(s_t, a_t\right)=\mathcal{I}\left(s_t\right)+\left(1-\mathcal{I}\left(s_t\right)\right)\left[  \mathbb{E}_{s_{t^{\prime}} \sim T\left(\cdot \mid s_t, a_t \right)} \mathcal{I}\left(s_{t^{\prime}}\right) \right]$$ 
As with the unsafe state indicator function, we can factorize the Safety Q-function into an \textit{ensemble of Safety Q-functions}, $\{Q_{\mathrm{safe,i}}\}_{i=0}^K$, each one trained for each type of failure, $Q_{\mathrm{safe}}\left(s_t, a_t\right)= \max\limits_{i} Q_{\mathrm{safe,i}}\left(s_t, a_t\right)$.

Training these Safety Q-functions in the real world would be dangerous, as the robot needs to experience and learn what actions may lead to unsafe states and when. 
Therefore, we pretrain an ensemble of Safety Q-functions in simulation, one for each type of unsafe transition. 
The ensemble of Safety Q-functions is pretrained in domain-randomized environments in the simulator OmniGibson~\cite{li2023behavior} in different scenarios, including articulated object interaction, rigid-body pick-and-place, and base navigation by sampling random and noise-corrupted task-related actions as generated by a motion planner.
The state representation consists of simulated pointclouds and robot proprioceptive information (for details of the network architecture, see Appendix~\ref{sec:app_details}).
Pointclouds provide a smaller sim2real gap than RGB images, allowing us to train our ensemble of Safety Q-functions in simulation and apply them zero-shot to safely explore in the real world.

\begin{figure*}[ht!]
\newcommand\fighione{2.15}
\newcommand\fighitwo{2.15}
\centering
\includegraphics[height=\fighitwo cm]{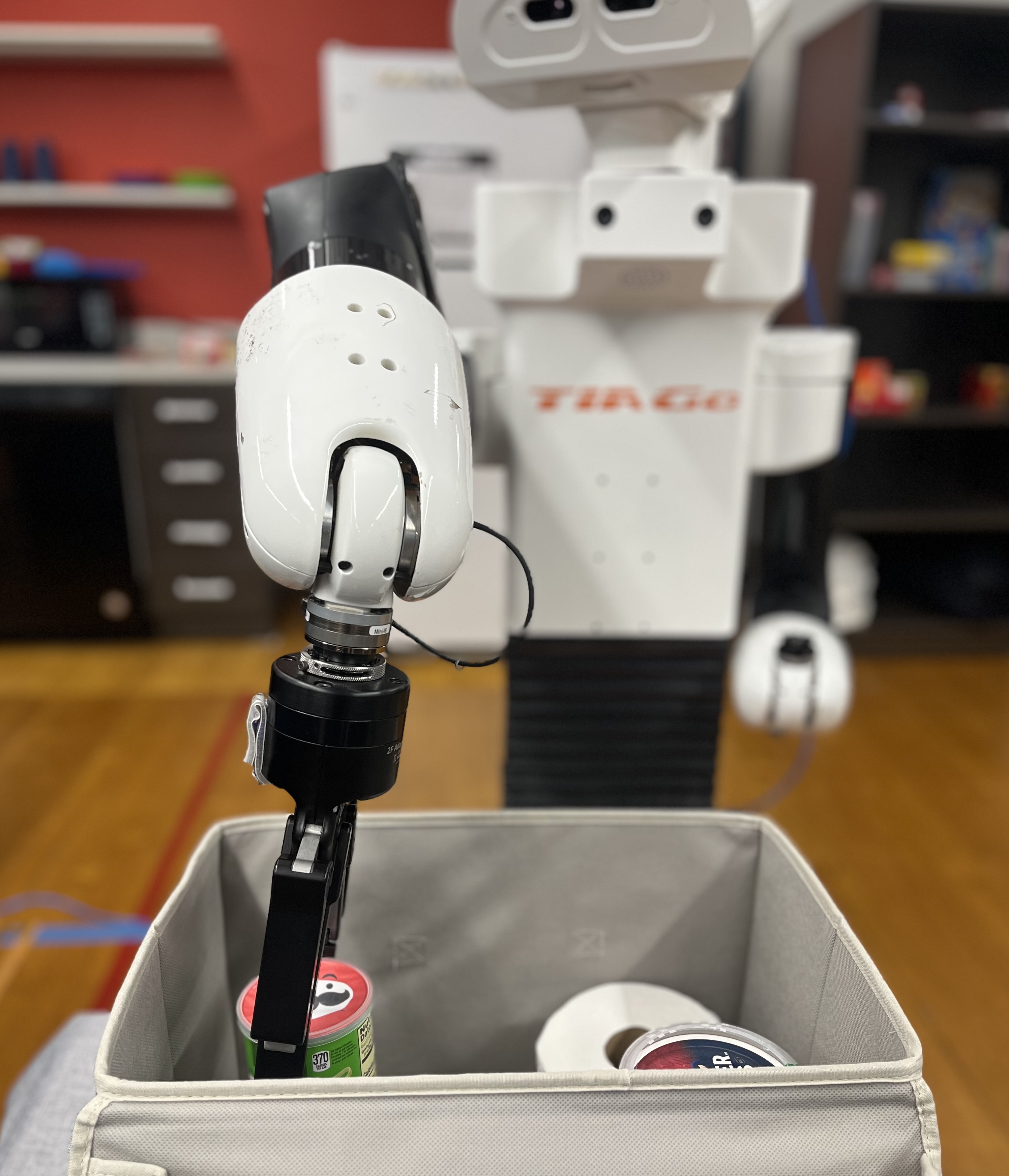}%
\hfill
\includegraphics[height=\fighitwo cm]{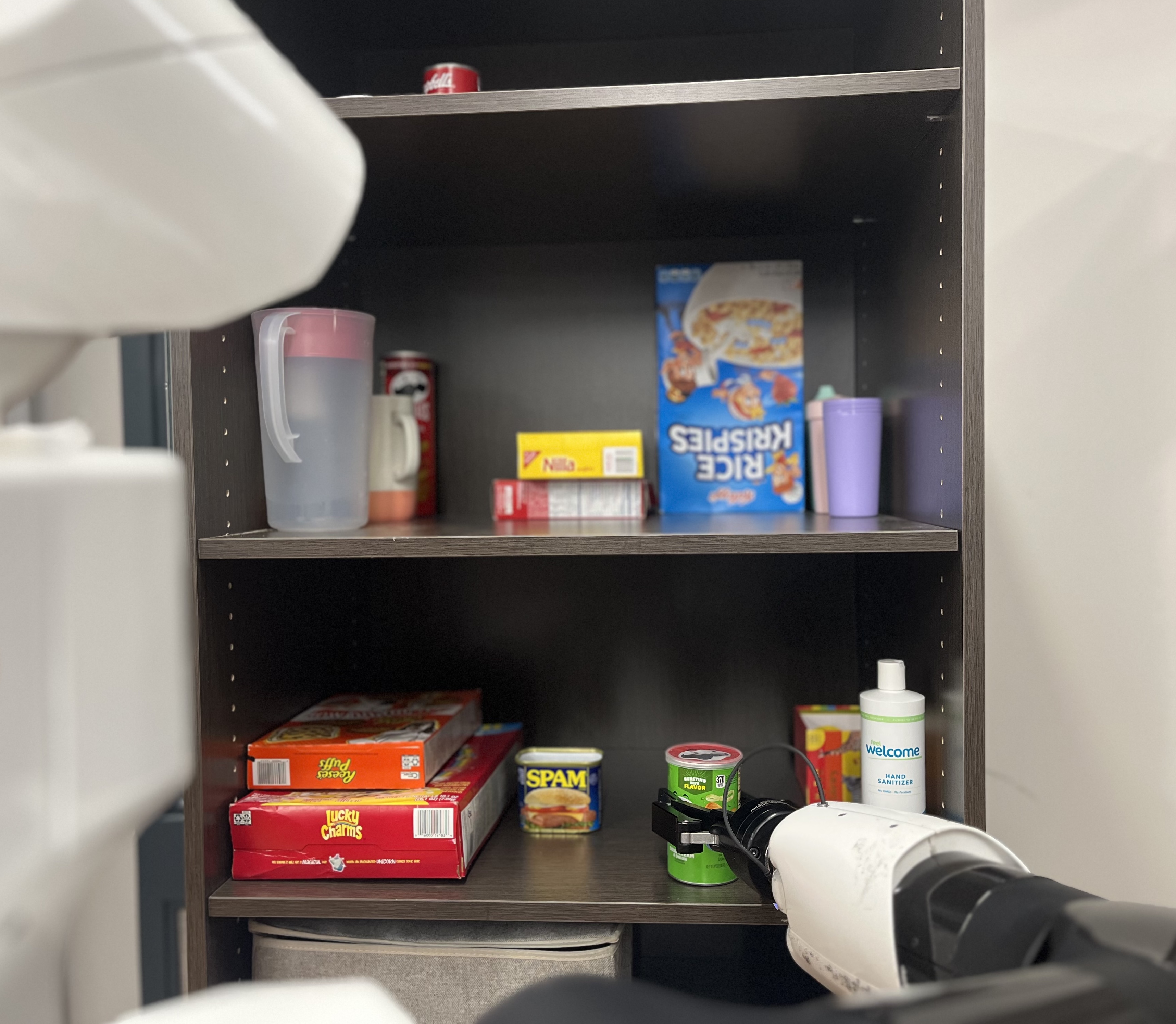}%
\hfill
\includegraphics[height=\fighitwo cm]{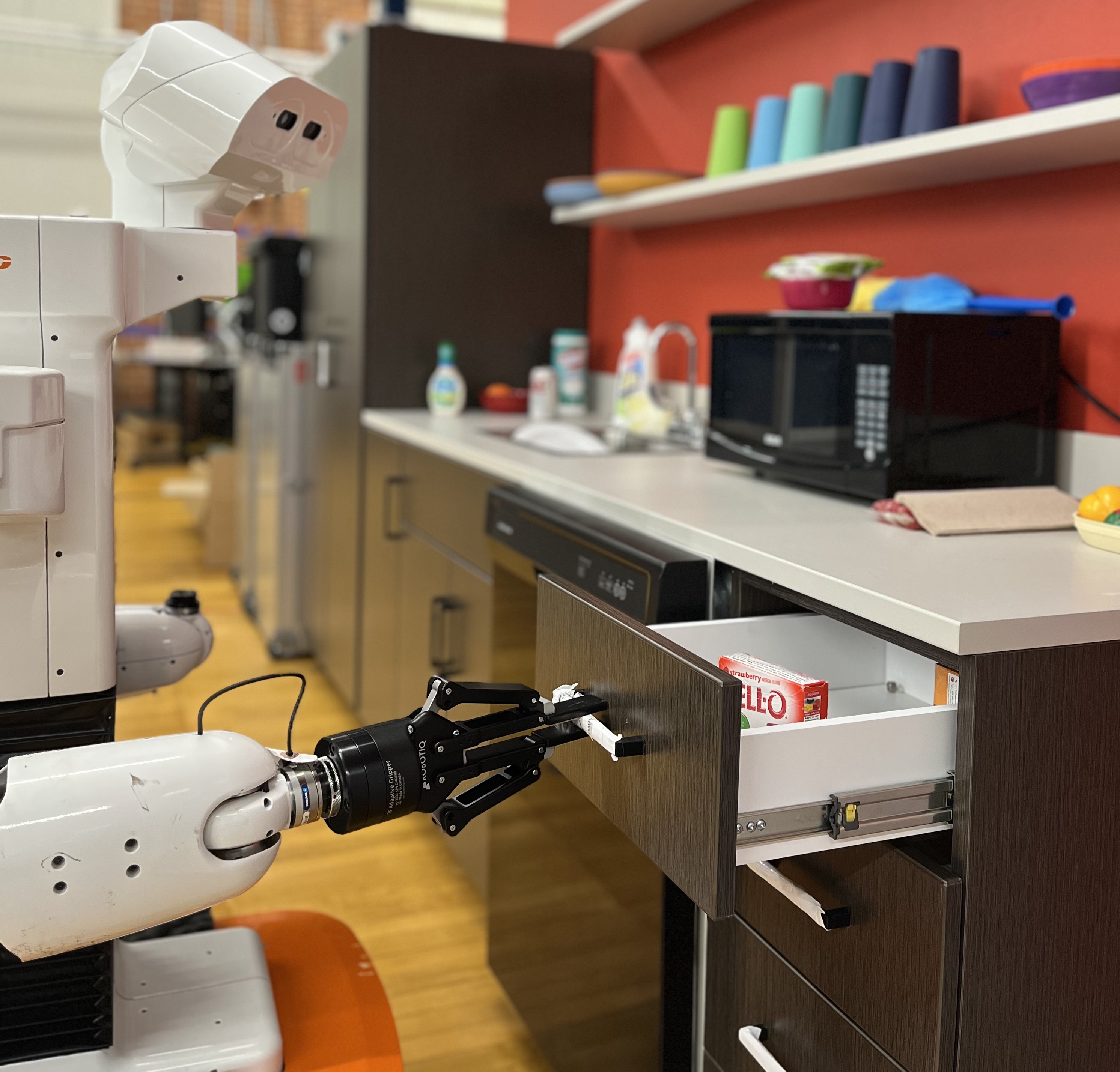}%
\hfill
\includegraphics[height=\fighione cm]{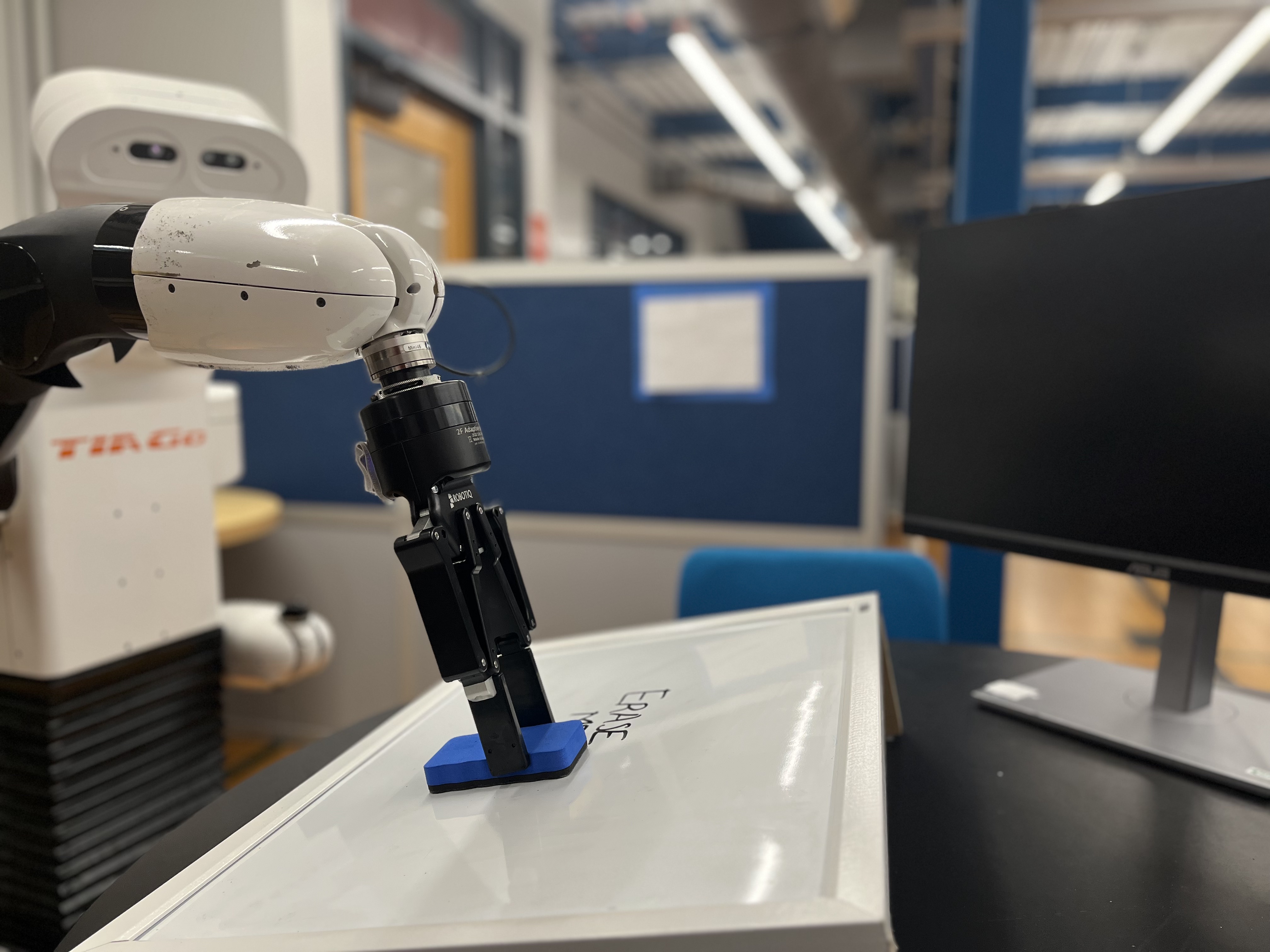}%
\hfill
\includegraphics[height=\fighitwo cm]{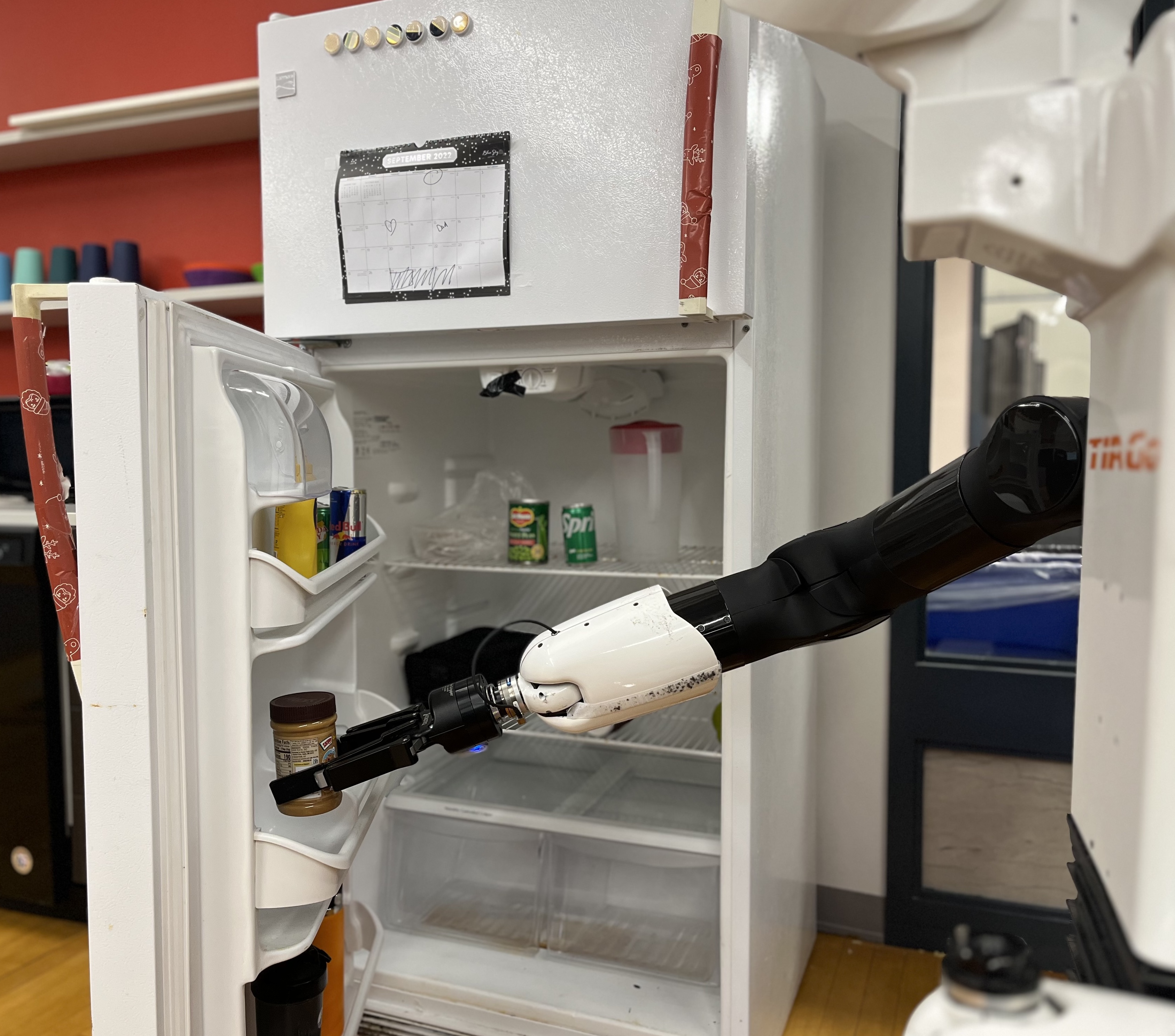}%
\hfill
\includegraphics[height=\fighione cm]{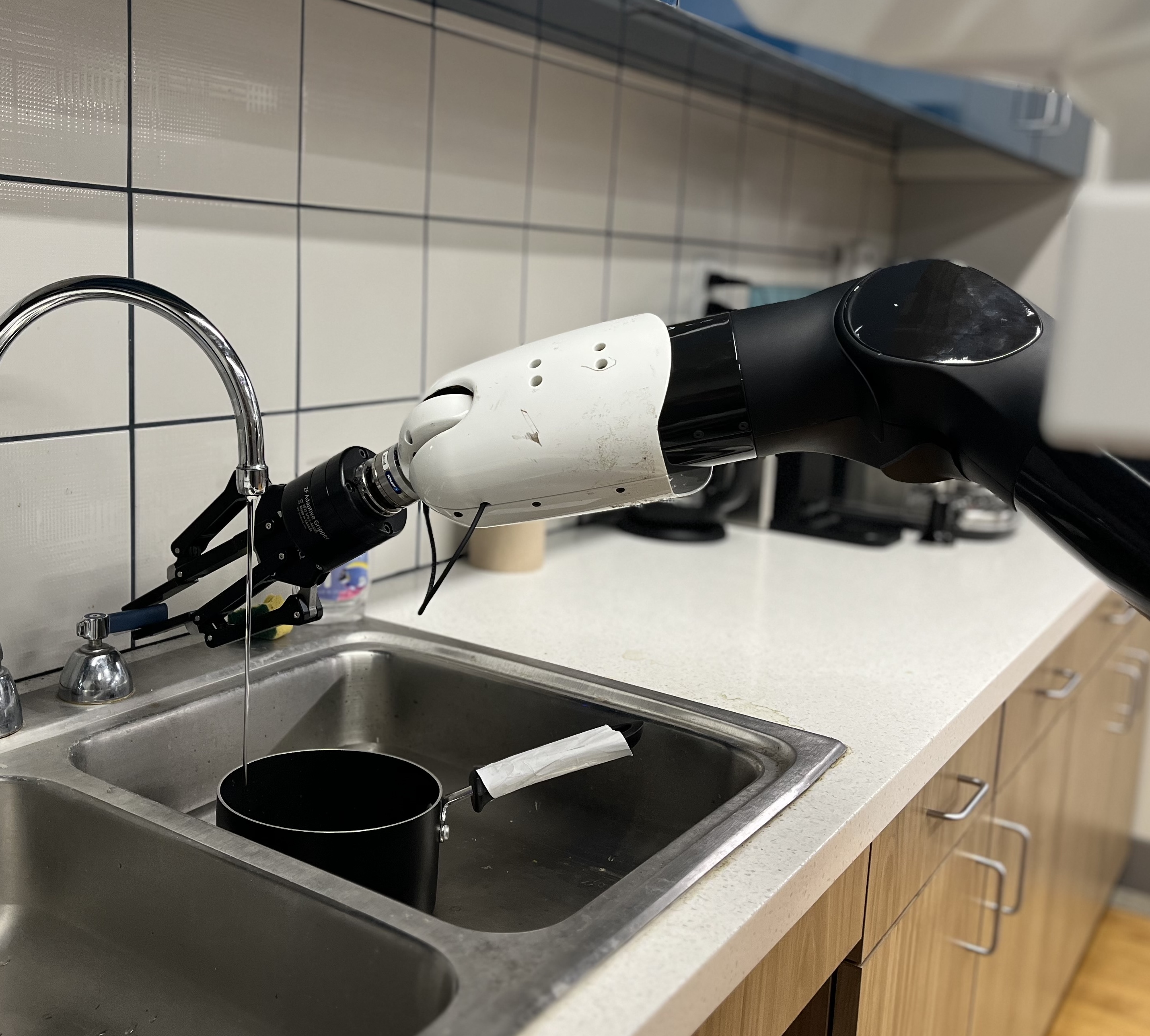}%
\hfill
\includegraphics[height=\fighione cm]{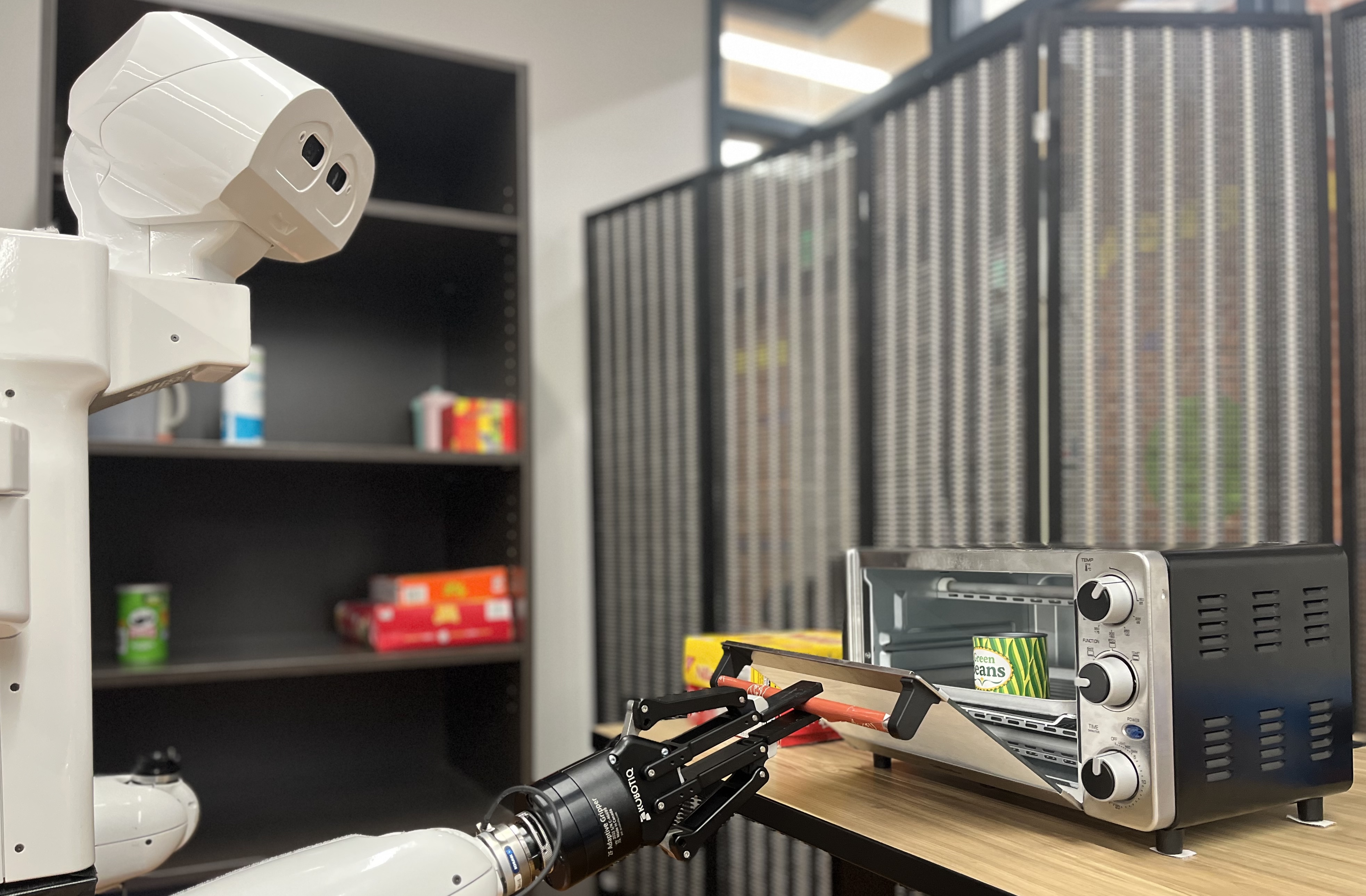}%

\caption{\textbf{Multi-Step Mobile Manipulation Evaluation Tasks.} \methodname{} is evaluated on seven complex multi-step tasks combining navigation and manipulation. From left to right: \texttt{boxing} an item, \texttt{shelving} an item, \texttt{store\_in\_drawer}, \texttt{erase\_whiteboard}, \texttt{refrigerating} an item, \texttt{fill\_pot}, \texttt{load\_oven}. The tasks involve risky contact-rich phases, grasping steps and the manipulation of constrained mechanisms. \methodname{} is able to adapt actions obtained from one single human demonstration safely and autonomously.}
\label{fig:tasks} 
\vspace*{-1em}
\end{figure*}

During the safe adaptation and exploration procedure, \methodname{} samples candidate actions around the human demonstrated motion.
For the grasping action, \methodname{} explores among a discrete set of options (3 grasps in our case) provided by a robot grasp generator~\cite{ten2017grasp}, first prioritizing the grasp closest to the human demonstrated one.
For continuous motion, \methodname{} explores action sequences from a Gaussian distribution with the mean given by the parsed human motion and variance adapted to encourage exploration. 
Sampled actions with the lowest $Q_\mathrm{safe}$ are selected and executed at each step, after which \methodname{} evaluates for segment completion based on the parsed semantic goal of the segment and a VLM.
If the segment has not been completed, the process repeats with a new set of actions around the human demonstrated one (see pseudocode in Alg.~\ref{algo:algo}).

\textit{Autonomous Exploration with Backtracking Mechanisms:} 
Thanks to the previous process, \methodname{} generates a safe exploratory behavior around the human-demonstrated motion that leads to adaptation. 
However, human monitoring would still be necessary to reset the robot and the environment and attempt new sequences of actions.
We aim at reducing this dependency in \methodname{}.
To that end, we implement a simple but effective \textit{backtracking} mechanism.
During safe exploration, if no sampled actions are safe from the current state (i.e., $Q_\mathrm{safe}(s, a_i) > \epsilon$ for all sampled $a_i$), the agent backtracks one step in the exploration: it executes the inverse to the last action taken, i.e., performs the opposite motion.
When interacting with objects (e.g., opening a door or drawer, moving a grasped object), this backtracking mechanism will bring the environment to the previous state, allowing \methodname{} to explore a different branch of the state-action space, verifying and executing actions until either the segment is complete or the maximum number of attempts is reached.

The safe and autonomous exploration from above allows the robot to find minor adaptations to the parsed human trajectories for each segment.
However, often, adaptation needs to be larger than a small change in motion, especially when it comes to adapting grasping strategies from a human to a robot.
When, after several iterations (50 actions in our case) of the single-step backtracking mechanism, \methodname{} exhausts all safe motion alternatives, it backtracks until it reaches a grasping segment, and selects a different grasp mode to explore.
We observe that this exploration of grasping modes combined with trajectory-level probing is critical to adapt multi-step human strategies into successful ones for the robot (see Fig.~\ref{fig:grasp_modes} for examples of grasping mode exploration). 

\subsection{Learning from Previous Successful Exploration}
\label{ss:tc}
Through our safe and autonomous exploratory process, \methodname{} adapts the motion and sequence of semantic changes initially parsed from the human video.
However, successful strategies need to be learned in order to prevent repeated exploration for the same task.
To that end, \methodname{} integrates a policy memory mechanism that biases exploration based on previous successes.
The policy memory biases the exploration (of grasp modes and of motion actions) from the human demonstrated to one that demonstrates success for the robot.
To that end, we then train an action prediction policy network that maps point clouds, $P$ and language description of the task, $l$ to actions, e.g., the grasping mode, $g \in SE(3)$, and sequence of post-grasp actions, $(a_0, \ldots, a_T)$, that led to success. 
The architecture for the action prediction policy network is composed by a PointNet~\cite{qi2017pointnet} encoder for the visual information, and a SentenceTransformer~\cite{reimers-2019-sentence-bert} for the task description, combined with an MLP head, and trained with geometric augmentations (rotations and translations) of the data from successful trials.
This model associates successful strategies to geometric and language information about the task, and allows \methodname{} to predict them from different viewing angles, a critical capability to leverage successful exploration for new instances of the same multi-step mobile manipulation tasks (see Fig.~\ref{fig:q3_policy_memory}). Furthermore, the model also helps reduce future explorations for the same task (see Exp. 3, Fig.~\ref{fig:q3_policy_memory}).

\section{Experimental Evaluation}
\label{sec:results}

\begin{figure*}[ht!]
\centering
\centering


\newcommand\fighithree{2.8}
\newcommand\fighifour{2.8}

\includegraphics[width=0.8\textwidth]{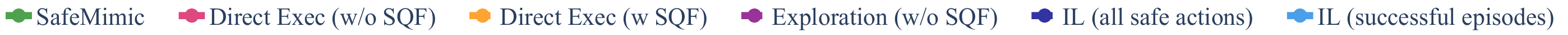}\\
\includegraphics[height=\fighifour cm]{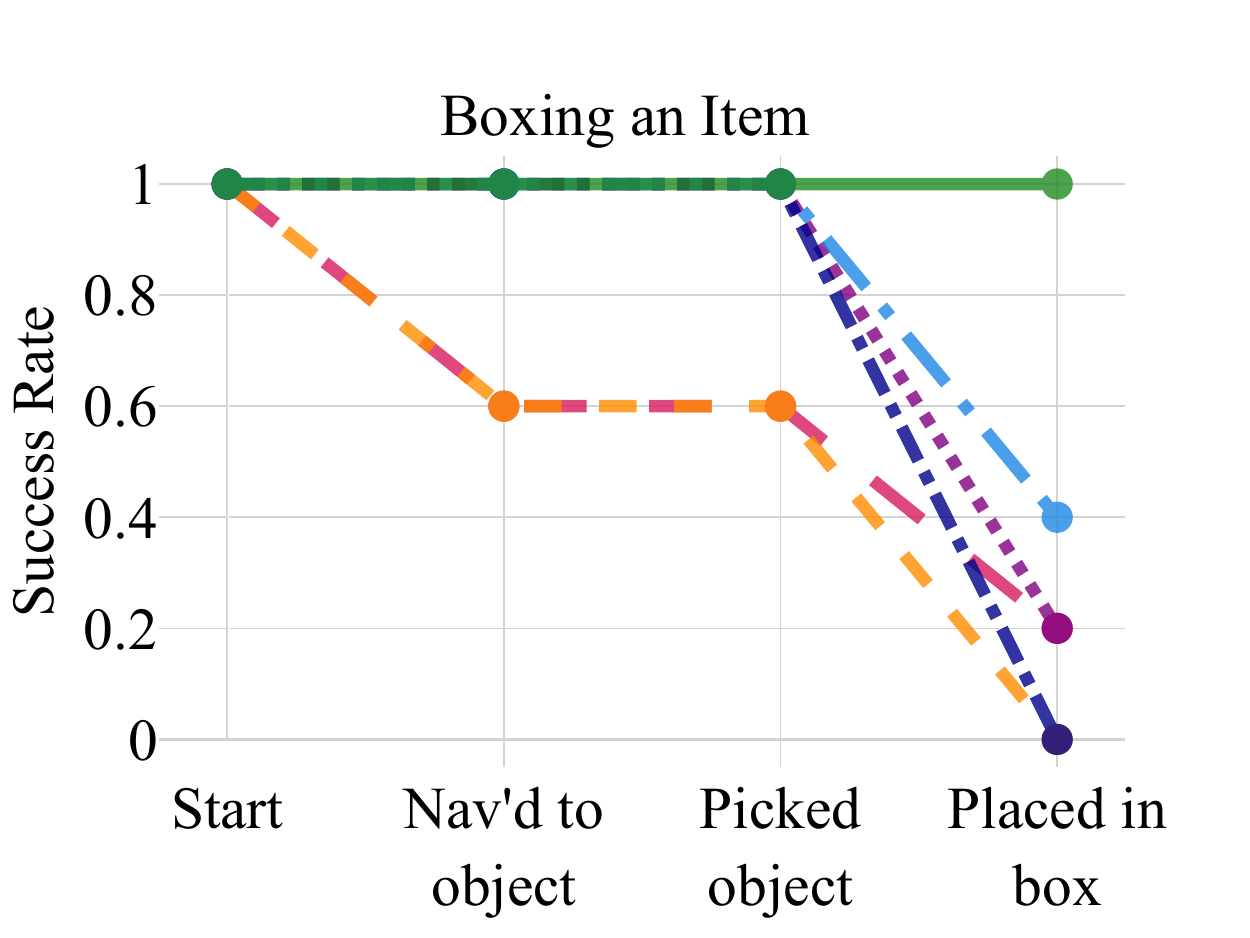}%
\hfill
\includegraphics[height=\fighifour cm]{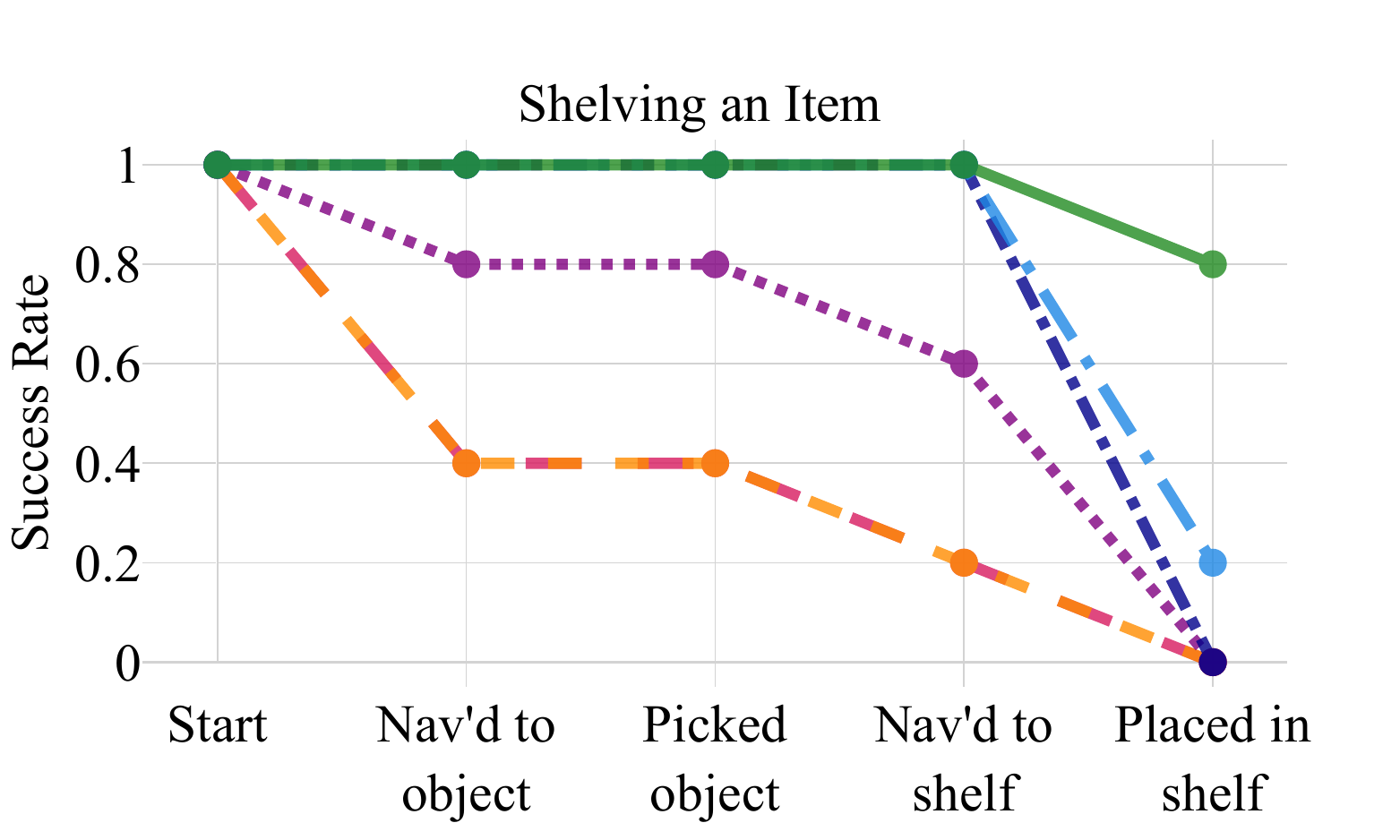}%
\hfill
\includegraphics[height=\fighithree cm]{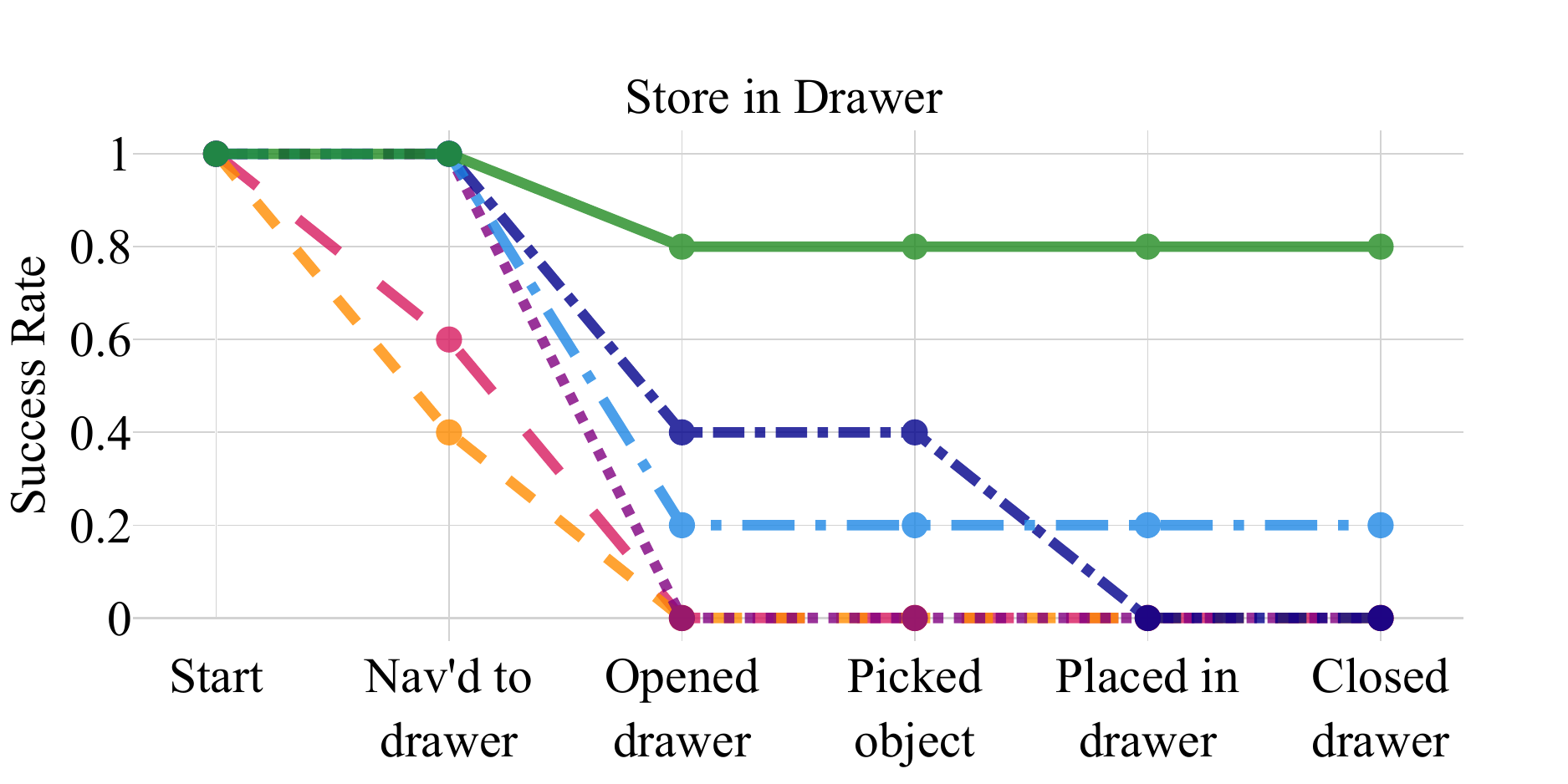}%
\hfill
\includegraphics[height=\fighifour cm]{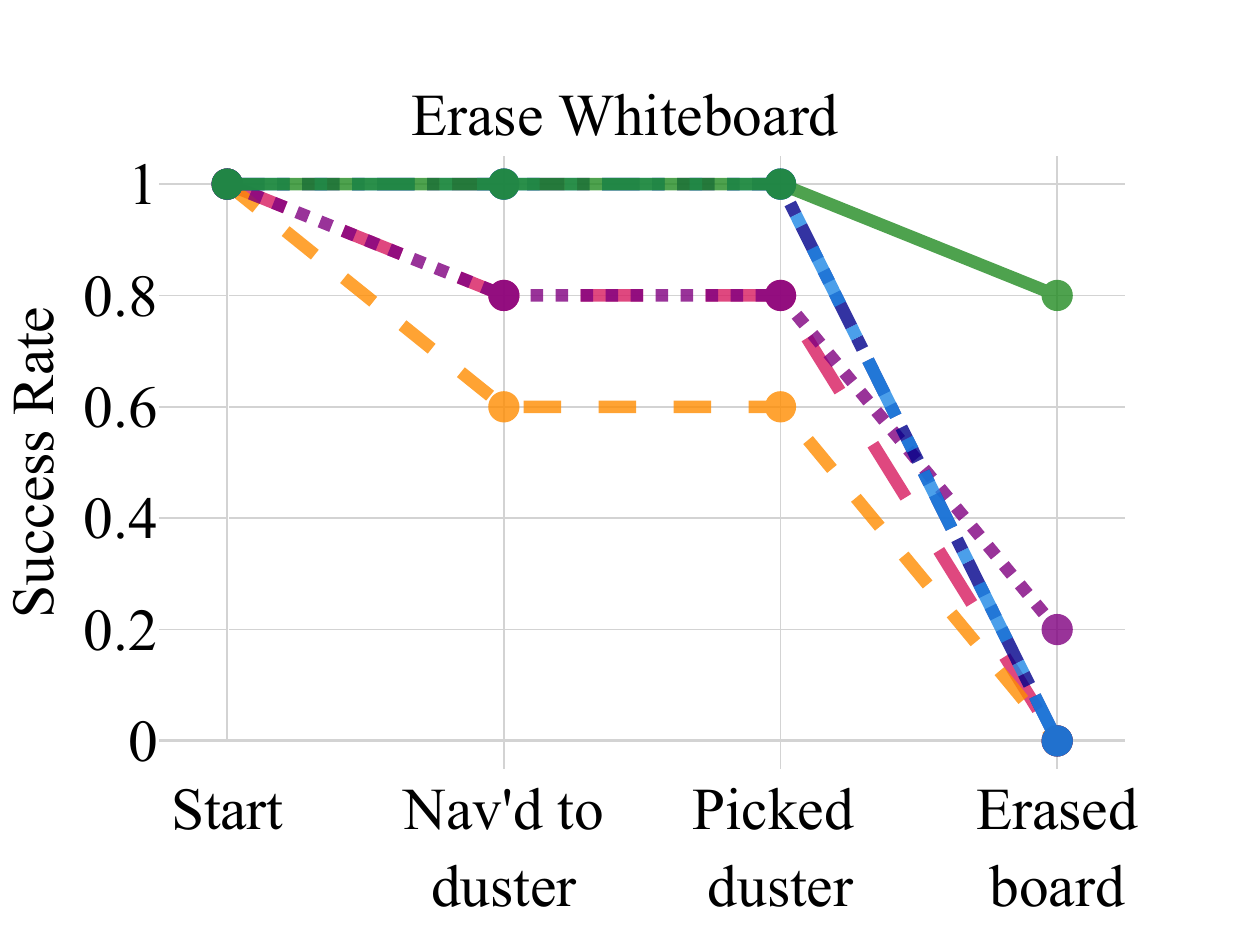}%
\hfill
\includegraphics[height=\fighifour cm]{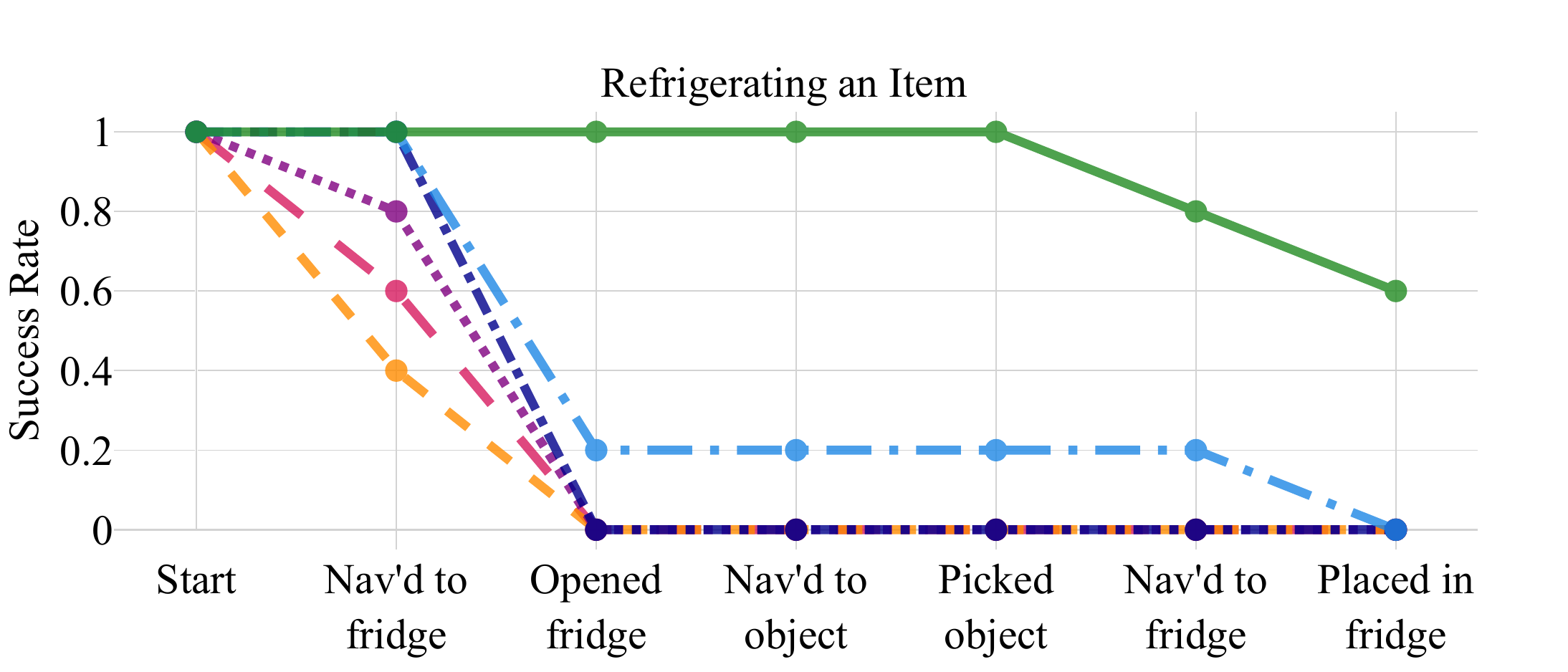}%
\hfill
\includegraphics[height=\fighifour cm]{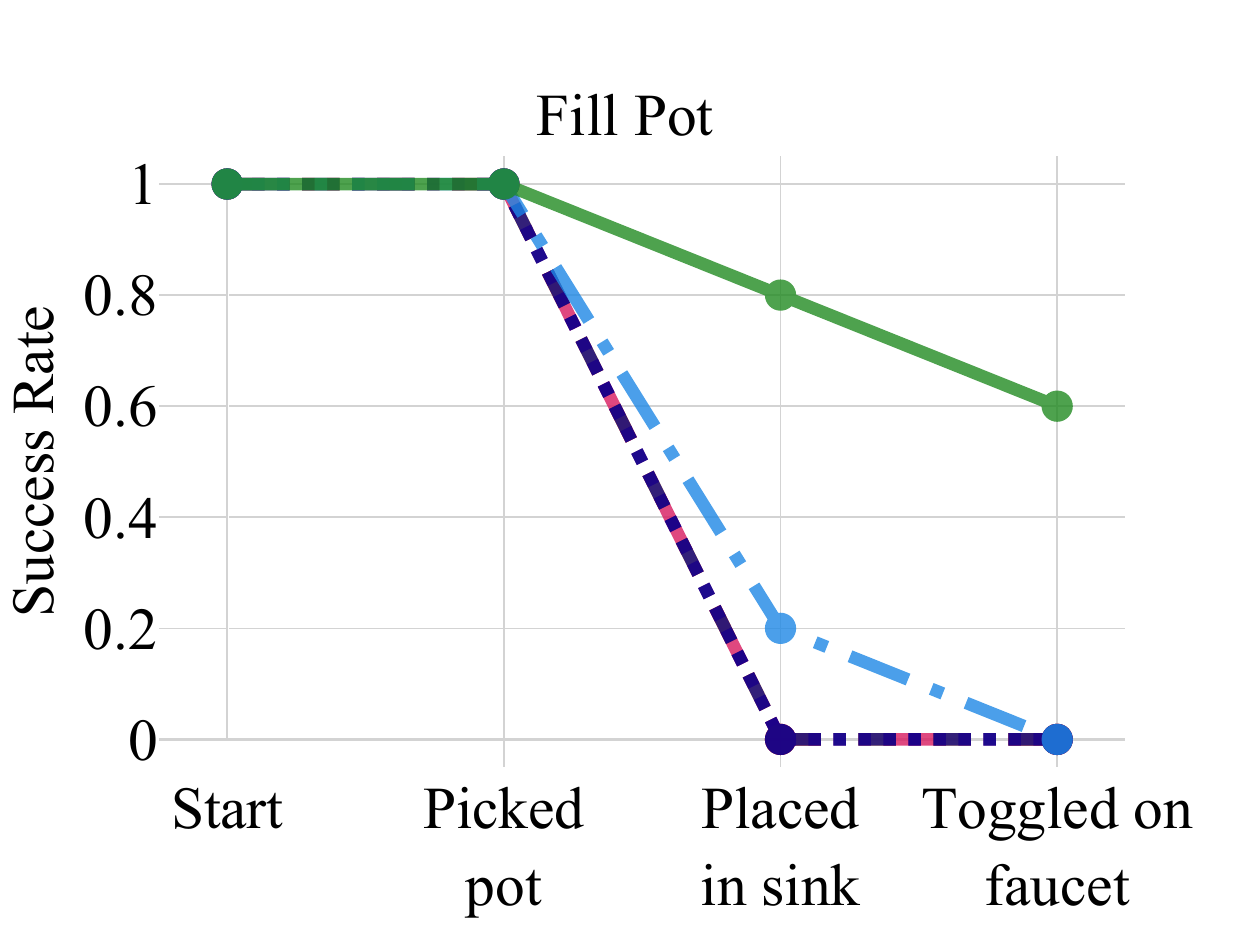}%
\hfill
\includegraphics[height=\fighithree cm]{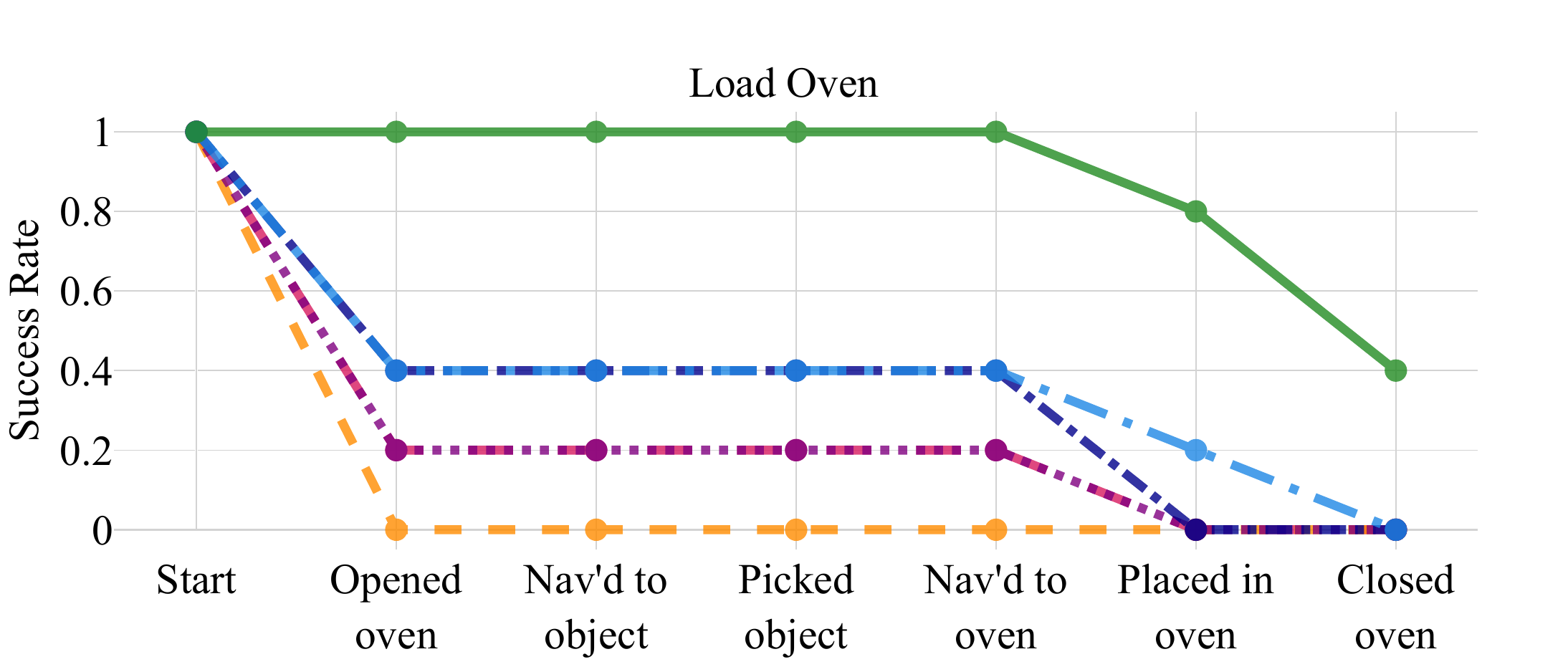}%


\caption{\textbf{Accumulated Success on Multi-Step Tasks}. Accumulated success rate at each stage of each of the seven evaluated multi-step mobile manipulation tasks, indicating the percentage of the five trials each method completed up to and including that segment. We compare \methodname{}'s task performance to five baselines: direct execution without safety Q-functions (SQFs), which requires human supervision, direct execution with SQFs, exploration without SQF, Imitation Learning (IL) with all safe actions in simulation and IL only with the successful episodes. Note that the IL baselines were evaluated \textbf{only} on ``Place'', ``Open'', and ``Close'' segments and were provided successful solutions for navigation and pick segments. Note as well that some lines overlap at the same rate. Across all tasks, \methodname{} significantly outperforms all baselines and achieves up to 100\% success in exploratory adaptation, indicating a superior ability to safely and autonomously refine the demonstrated behaviors.}
\label{fig:success_rate} 
\vspace*{-1em}
\end{figure*}

We evaluate \methodname{} in 7 challenging multi-step mobile manipulation tasks demonstrated by humans. The tasks all consist of multiple stages and require navigation, rigid-body pick-and-place, articulated object interaction, and contact-rich control, all while avoiding common failure modes, including collisions, joint limit violations, force-torque limits, and grasp loss. We use these tasks to evaluate \methodname{}'s task performance, safety, generalization, and robustness to different users and environments.
In the following, we briefly explain each task.

\begin{itemize}[leftmargin=*]
    \item \texttt{boxing} an item:
    Pick the object from the table and place it in a box, requiring a top-down grasp by the robot to avoid collision.  
    \item \texttt{shelving} an item: Pick an object from the table and store it on a shelf above the sink, or in a bookshelf, as demonstrated. This task requires differentiating the human's semantic goal, and avoiding collisions and adapting grasps for successful placement. 
    \item \texttt{store\_in\_drawer}:
    Open the drawer, pick the object from the table, store the object in the drawer, and close the drawer. This task consists of several segments, including an articulated object interaction that typically necessitates adapting human-demonstrated grasps to avoid joint limit violations. 
    \item \texttt{erase\_whiteboard}: Grasp an eraser and erase the writings on a whiteboard. This task requires contact-rich control, demanding that the robot avoid force-torque limits in particular. 
    \item \texttt{refrigerating} an item:
    Open the refrigerator, pick the object from the table, store it in the refrigerator. This task requires adapting the grasp poses and demonstrated trajectories to open the large, heavy refrigerator door and place the object in a constrained space.    
    \item \texttt{fill\_pot}:
    Pick a saucepan, place it in the sink, and toggle on the faucet. This task necessitates adapting the commonly used human grasp to one that the robot can use to lower the pot into the sink.
    \item \texttt{load\_oven}:
     Open the oven, pick the object, place it in the oven, and close the oven. This task consists of several segments, and requires a specific side-grasp to successfully place the food in the small oven.
\end{itemize}

In all the experiments, we use a PAL-Robotics Tiago++ mobile manipulator. We control one arm's end-effector pose using IK control~\cite{beeson2015trac} and the base using relative position and yaw commands. For perception, we use an Orbbec Astra S RGB-D camera mounted on Tiago++'s head both to observe humans and as input to the safety Q-functions, and an ATI mini45 force-torque sensor mounted on the wrist to detect and predict excessive force-torque violations. 
While \methodname{} is generic and can include many possible failure modes, we consider the following in this work: arm collisions, base collisions, joint limit violations, force-torque limits, grasp
loss, and dropping objects.

Throughout our experiments, we compare \methodname{} against the following baselines: \texttt{Direct Execution (w/o SQF)} directly executes the actions obtained from the video parsing module and does not verify those actions using the Safety Q-Functions (SQFs). \texttt{Direct Execution (with SQF)} directly executes the human actions as well but verifies the actions using SQFs, avoiding unsafe robot actions. \texttt{Exploration (w/o SQF)} performs exploration starting from the human tracking actions but does not use SQFs. This baseline is \methodname{} without the use of SQFs. 
We also evaluate if the data generated to train our safety Q-functions would suffice for training task policies: we include \texttt{Imitation Learning (IL)} baselines based on a BC-RNN Behavior Cloning policy with a PointNet encoder trained on two types of datasets: \texttt{IL (all safe actions)} in which the policy is trained on all simulation data where the state-action pairs did not lead to safety criteria being violated, and \texttt{IL (successful episodes)} which is trained only on the subset of successful task executions in the simulator. Since the data collection for navigation is task-agnostic (random base commands), we can not train a task-oriented IL policy on that data. \arpit{Furthermore, since SafeMimic performs picking action based on a grasp generator, we also don't train an IL policy for picking. Instead, we use the successful navigation and picking segments from the \methodname{} trials for this baseline. Hence, navigation to an object and picking always succeeds and we focus the comparison on the manipulation segments of "Place”, “Open”, and “Close” for this baseline. Note that we trained separate IL policies for each segment.}

\subsection*{Experiments and Results}
In our experiments, we aim to answer four questions: \\
\textbf{Q1) Does \methodname{} enable a robot to successfully complete a multi-step mobile manipulation task from a third-person demonstration?} 
To evaluate \methodname{}'s task performance, we measure its ability to successfully imitate and adapt demonstrations of the aforementioned tasks. For each task, \methodname{} and the baselines begin with the initial trajectory obtained by tracking the human pose, broken into semantic segments identified by the parsing module. Each task is attempted five times, and success rates are reported at each stage of each task.
The success rate for each segment (e.g., ``Navigated to object," ``Picked object") reflects the proportion of trials successfully completed up to and including that segment. \methodname{}'s real-world fine-tuning takes on average five minutes per navigation segment and 15 minutes per manipulation segment, and success is confirmed by prompting a VLM with the observation and semantic goal. 

Fig.~\ref{fig:success_rate} depicts the results of our analysis. We observe that \methodname{} achieves a minimum of 40\% final success rate over the seven tasks, significantly outperforming all baselines. The \texttt{Direct Execution} baseline achieves 0\% final success rate on all the seven tasks, demonstrating the need for exploration in order to effectively adapt the human demonstrations to the robot's morphology. Although \texttt{Direct Execution (w/o SQF)} and \texttt{Direct Execution (with SQF)} achieve similar success rates, \texttt{Direct Execution (w/o SQF)} required 82\% more human interventions due to potentially unsafe actions as compared to \texttt{Direct Execution (with SQF)}, showing the importance of our learned SQFs. Interestingly, we observe that \texttt{Direct Execution (w/o SQF)} achieves higher success than \texttt{Direct Execution (with SQF)} in certain segments due to false positives of the SQFs. 
While the imitation-learning baselines demonstrate some successes, they fail to reliably perform the tasks, indicating that while the small amount of noisy data we generate in simulation is sufficient to train \methodname{}'s SQFs, training robust imitation learning policies has a higher data quantity and quality demands. 
\begin{figure}[t!]
\centering
\newcommand\fighifive{3.3}
\newcommand\fighisix{3.5}
\centering
\includegraphics[height=\fighifive cm]{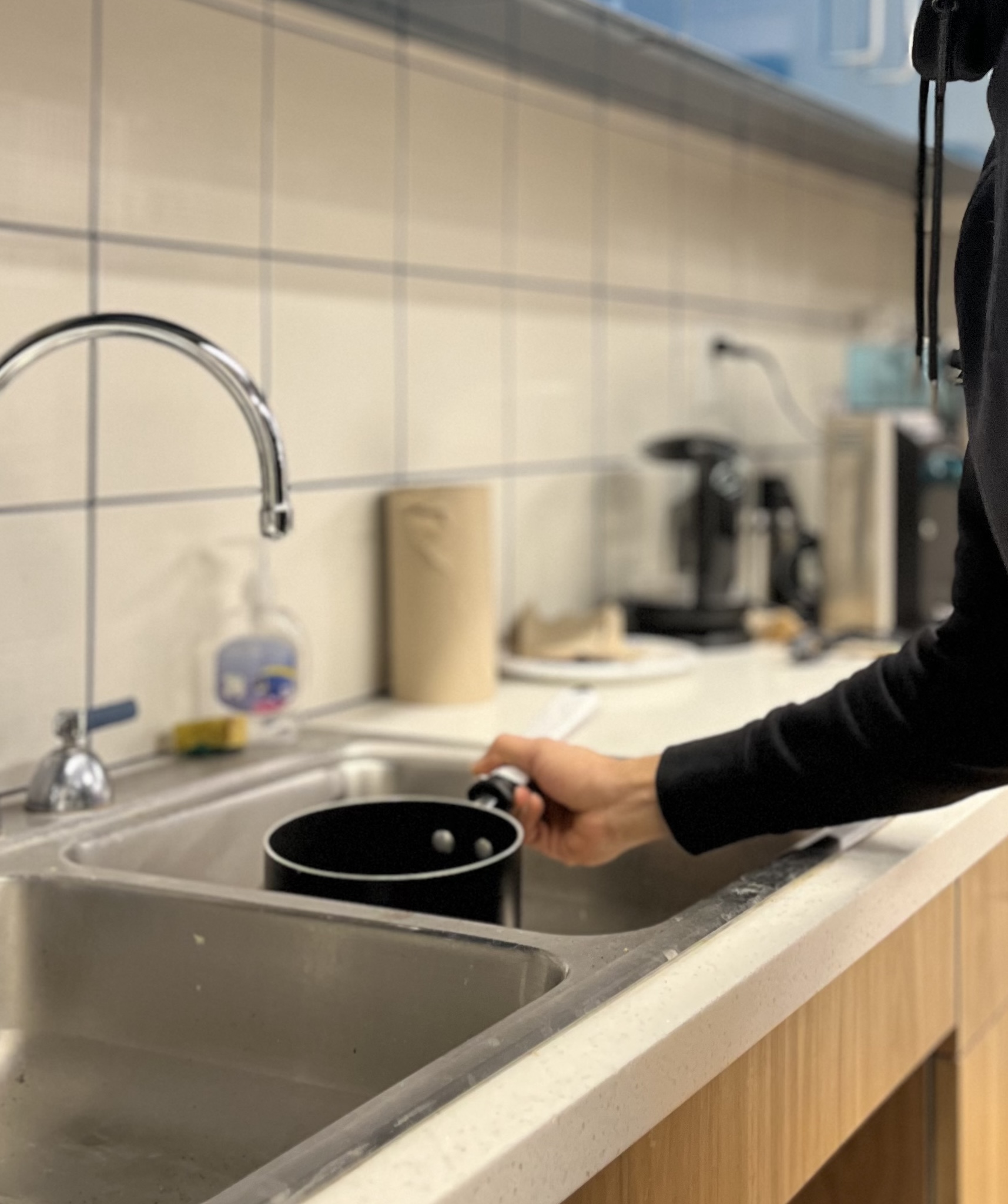}%
\hfill
\includegraphics[height=\fighifive cm,cfbox=red 1pt 0pt]{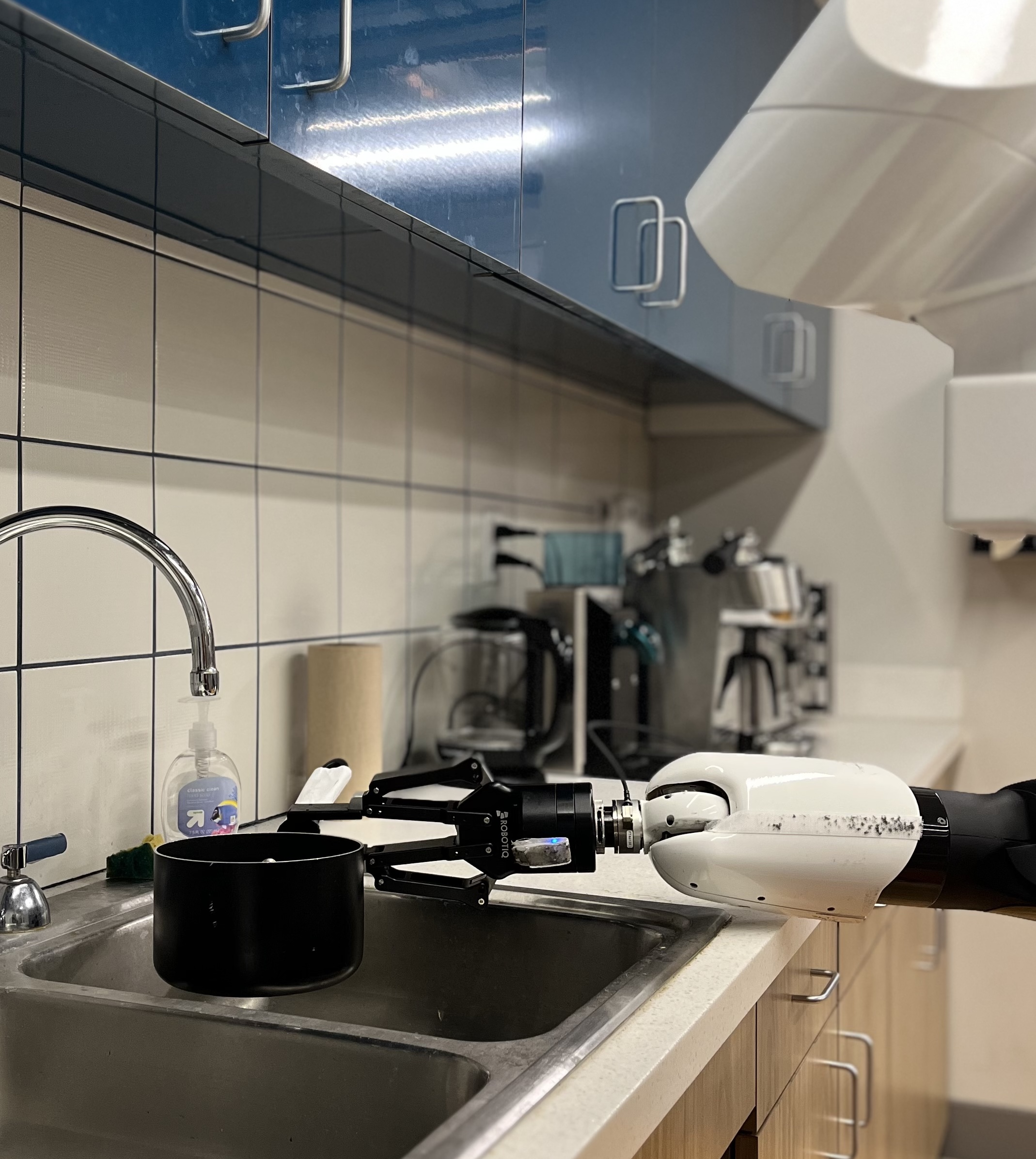}%
\hfill
\includegraphics[height=\fighifive cm,cfbox=green 1pt 0pt]{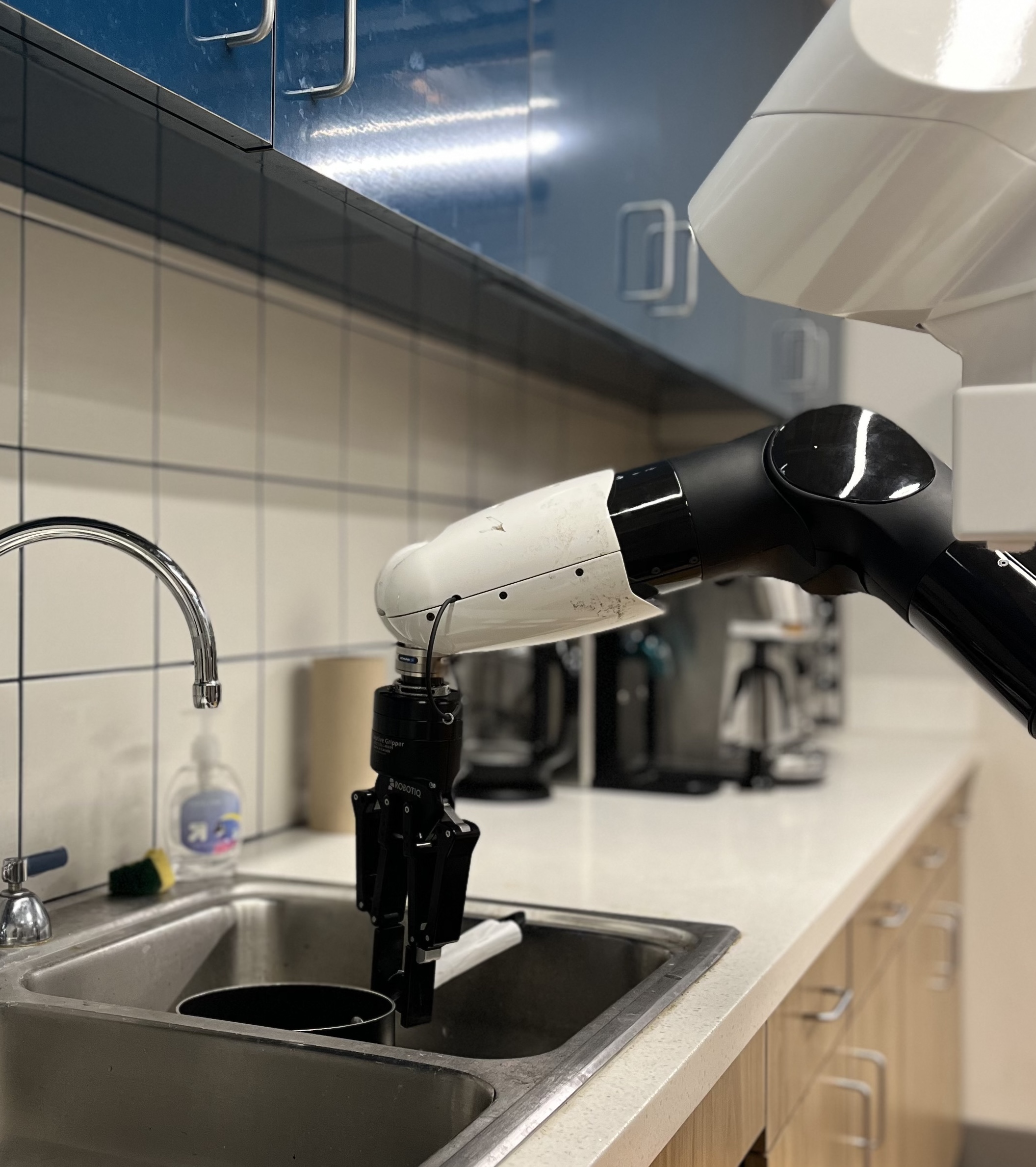}%
\\\vspace{5pt}
\includegraphics[height=\fighisix cm]{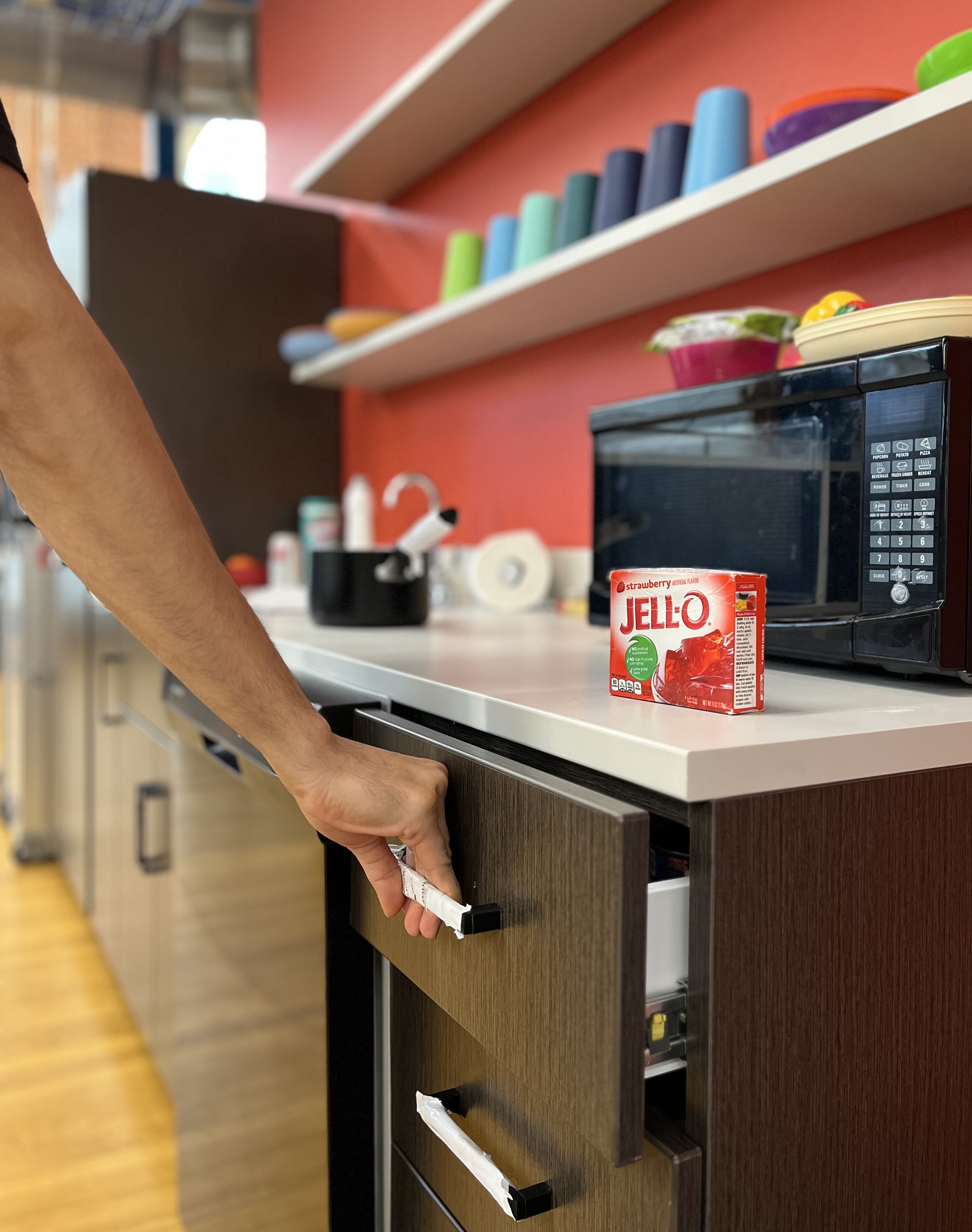}%
\hfill
\includegraphics[height=\fighisix cm,cfbox=red 1pt 0pt, trim={10cm 0 5cm 0},clip]{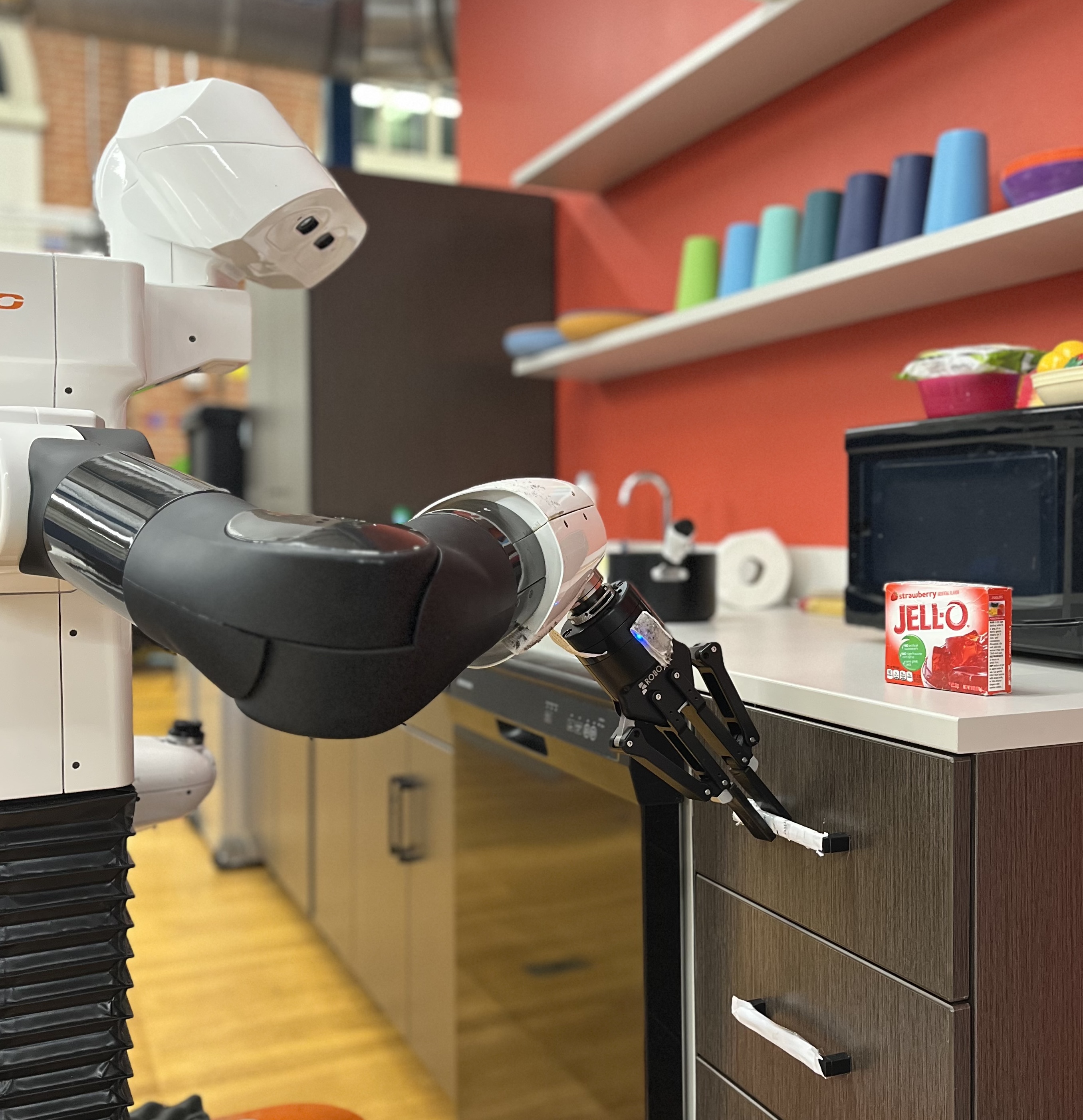}%
\hfill
\includegraphics[height=\fighisix cm,cfbox=green 1pt 0pt, trim={10cm 0 5cm 0},clip]{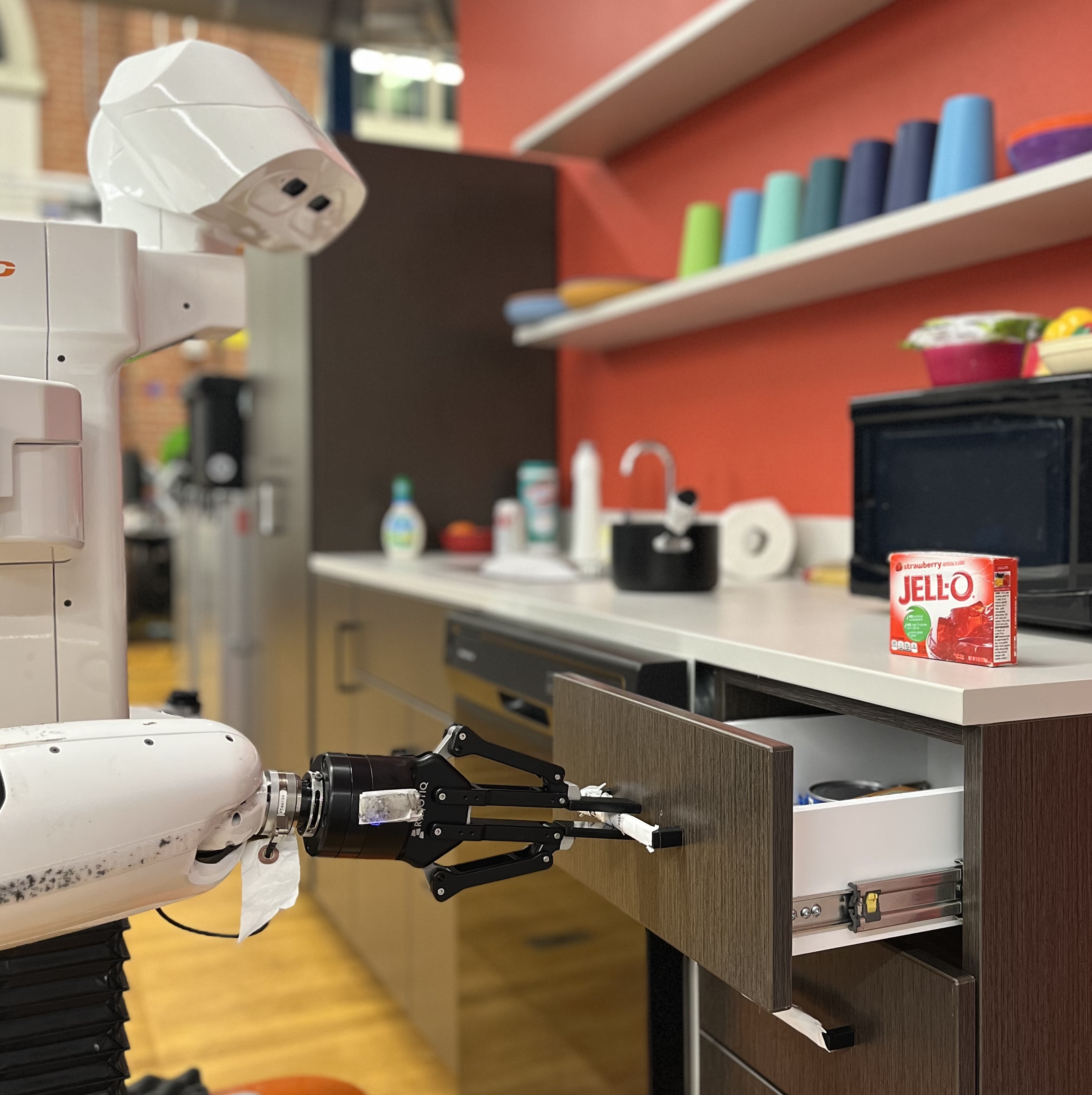}%
\caption{\textbf{Grasping mode adaptation.} Two examples (\textit{top and bottom rows}) of \methodname{}'s grasping mode adaptation. \textit{Left column:} human demonstrated grasp. \textit{Middle column:} robot failing when attempting the task by matching the human grasp. \textit{Right column:} robot succeeding in the task through \methodname{}'s grasp adaptation.
In 6 out of 7 tasks, \methodname{}'s grasp adaptation is critical to overcome human-robot embodiment differences and successfully imitate the demonstration.}
\label{fig:grasp_modes} 
\vspace{-2em}
\end{figure}

Grasp mode adaptation proved to be essential to \methodname{}'s success. For all but the \texttt{erase\_whiteboard} task, \methodname{}  adapted the grasp mode for one or more segments in order to succeed. 
Fig.~\ref{fig:grasp_modes} depicts two such examples: for the \texttt{fill\_pot} task, the robot is unable to place the pot in the sink with the human-like grasp, as the arm is predicted to collide with the edges of the sink as it attempts to place the pot. Hence, the robot backtracks to the start of the pick segment, samples and explores a top-down grasp, and successfully places the pot in the sink. Similarly, for the \texttt{store\_in\_drawer} task, the human-like grasp leads to joint limits being reached and so the robot explores and adapts its grasp to successfully open the drawer.

\begin{figure}[t!]
\centering
\newcommand\fighifive{3.67}
\newcommand\fighisix{3.67}
\centering
\includegraphics[height=\fighifive cm]{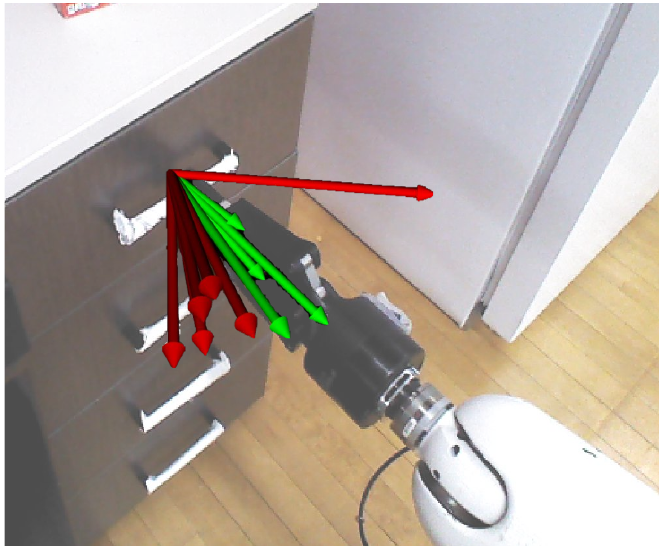}%
\includegraphics[height=\fighifive cm]{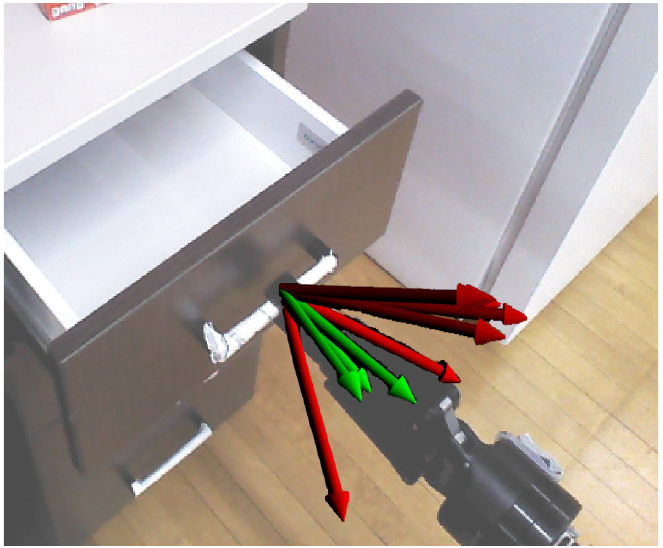}%
\hfill
\\\vspace{1pt}
\includegraphics[height=\fighisix cm]{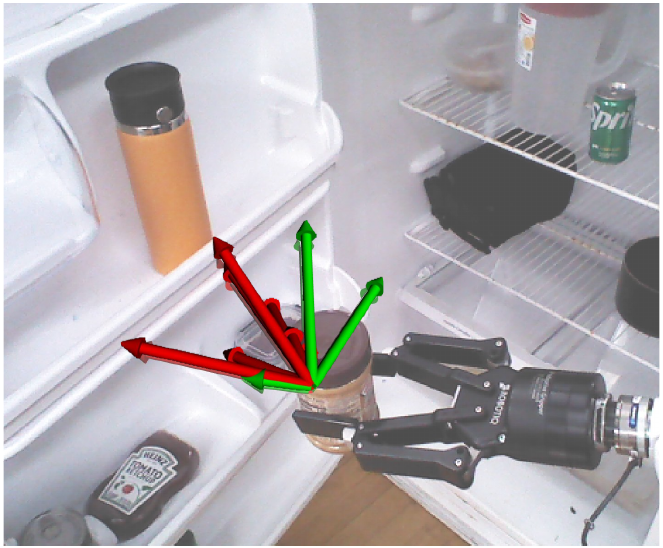}%
\includegraphics[height=\fighisix cm]{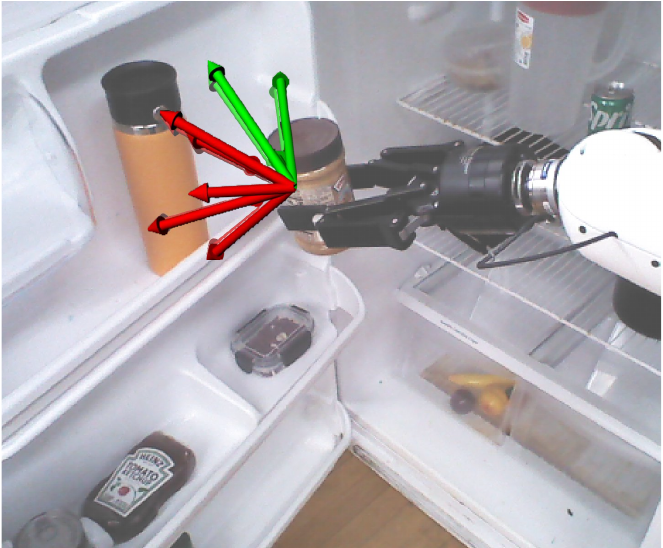}%
\caption{\textbf{Safe exploration with Safety Q-function predictions.} Examples of predictions of the Safety Q-function (SQF) for two tasks: the opening drawer segment in \texttt{store\_in\_drawer} (top-row) and placing a bottle in the fridge segment in \texttt{refrigerating} (bottom-row). Red arrows indicate predicted unsafe actions ($Q_{\mathrm{safe}} > \epsilon$) and green arrows show predicted safe actions. We observe that the simulation-trained SQFs reliably predict unsafe actions such as trying to open the drawer by moving in the wrong direction or trying to place an object in the fridge by colliding with the orange bottle. This enables \methodname{} to safely explore and adapt demonstrations in the real world.}
\label{fig:SQF_preds} 
\vspace*{-2em}
\end{figure}

\textbf{Q2) How effectively does \methodname{} reduce safety-critical failures?} To assess \methodname{}'s effectiveness for safe exploration, we compare the number of unsafe actions that happen during exploration with \methodname{} to the baselines. A human monitors the execution of \methodname{} and the baselines and flags unsafe actions. For each method, we measure the percentage of unsafe actions (\texttt{unsafe action rate}) as the ratio of all unsafe actions to the total number of actions the robot executed (over all trials for the seven tasks).

As expected, we observe that methods that do not use SQFs face the highest unsafe action rates. 
The \texttt{Direct Execution (without safety Q-functions)} generates 13.4\% unsafe actions and incurs safety violations in nearly every task, commonly colliding during both navigation and manipulation segments or reaching force limits measured by the FT sensor. 
Exploration alone (\texttt{Exploration without SQF}) similarly results in 14.2\% unsafe actions, demonstrating the critical need for safety during exploration absent in current frameworks for learning from human video. 
Both the IL baselines, \texttt{IL (all safe actions)} and \texttt{IL (successful episodes)} observe slightly lower unsafe action rates at 10.8\% and 9.5\% respectively, but still not low enough for safe deployment and adaptation. 
The inclusion of the SQFs in (\texttt{Direct Execution with SQFs}) and \methodname{} results in an unsafe action rate of 0.5 \% and 0.6\%, reducing the number of safety violations by 13.6\%, demonstrating the efficacy of the SQFs. Fig.~\ref{fig:SQF_preds} includes examples of predictions of safe and unsafe action samples during \texttt{store\_in\_drawer} and \texttt{refrigerate\_item}. It can be observed that actions that do not align with the drawer's kinematic constraints or that would lead to a collision during placing are correctly predicted to be unsafe by \methodname{}'s SQFs. 

\begin{figure}[t]
    \centering
    \vspace*{-2em}
    \includegraphics[width=0.95\columnwidth]{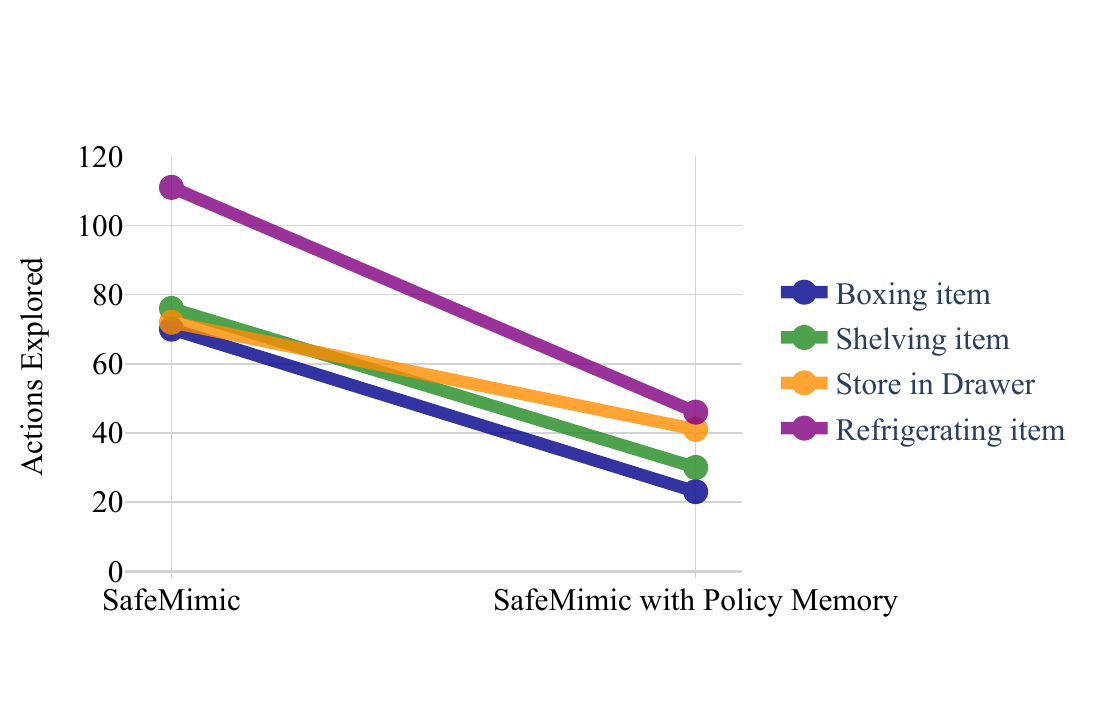}
    \vspace*{-2em}
    \caption{\textbf{Exploration Reduction with Policy Memory.} Number of actions explored by \methodname{} with (right) and without policy memory (left).
    Successful attempts from an initial exploration are recorded and use to train the policy memory.
    Object poses are varied relative to the robot before re-evaluating \methodname{} with the learned policy memory on several task segments. 
    The number of exploratory actions significantly reduces in all tasks, up to 67\%, demonstrating that learning from prior successes is critical to increase efficiency in mobile manipulation tasks. 
    }
    \label{fig:q3_policy_memory} 
    \vspace*{-1em}
\end{figure}

\textbf{Q3) Can \methodname{} help robots learn task-specific actions to reduce exploration in future attempts?}
\methodname{} enables safe and autonomous learning of a demonstrated task, and the policy memory module (see Sec.~\ref{sec:method}.D, Fig.~\ref{fig:method_diagram}, \textit{bottom right}) allows the agent to associate solutions to the task for future attempts. 
To evaluate this capability, we applied \methodname{} on several tasks and trained a policy memory module with successes. The memory module was trained to predict grasps for the pick segment of the tasks \texttt{boxing, shelving, refrigerate\_item}, opening segments of tasks \texttt{refrigerate\_item, store\_in\_drawer} and trained to predict the full action trajectory for the drawer opening segment of \texttt{store\_in\_drawer}. During evaluation, the objects to be picked and their poses were randomized, and due to the inherent stochasticity in the exploration phase of the navigation segments, the viewpoints are also variable during the evaluation (a common problem in mobile manipulation tasks). Three evaluation trials were performed for each task after initial refinement attempts during the first experiment. Fig.~\ref{fig:method_diagram} \textit{(bottom-right)}  depicts two different viewpoints of the drawer (and the corresponding action prediction) that the robot sees after completing the navigate to drawer segment. 
For the segments that are not learned, we reuse the actions obtained from the human video parsing module.
Fig.~\ref{fig:q3_policy_memory} depicts the results of our experiments. For the placing tasks, learning to predict successful grasping modes from previous adaptations dramatically reduces the number of waypoints explored during subsequent executions. In the case of \texttt{store\_in\_drawer}, the policy memory module enables the robot to manipulate the drawer successfully in future attempts, further reducing action exploration.\newline

\textbf{Q4) Is \methodname{} sensitive to demonstrations provided by different humans or in different environments?}
To further validate the robustness of \methodname{}, we conducted experiments with different users and in different environments. Three users provided demonstrations for the \texttt{shelving} task in the main environment used for previous experiments to assess whether \methodname{} performed similarly across human demonstrators. Each human was free to perform the task as they naturally would, resulting in potentially different grasping strategies and trajectories, as depicted in Fig.~\ref{fig:humansenvs}. For all three demonstrators, \methodname{} was able to successfully recover, parse and translate the demonstrated behavior. Interestingly, the robot had to adapt the grasping strategy for each of the three provided demonstrations, as all humans preferred a top-down grasp but such a grasp did not allow the robot to place the object in the shelf.
We also tested one human demonstrator across three environments on the \texttt{shelving} task to ensure \methodname{} is robust to different objects and room layouts. We performed 3 trials with the robot. \methodname{} achieved 100\%, 100\% and 66\% success in each of the three environments respectively, demonstrating the method's generalization capabilities. 

\section{Limitations and Future Work} 
\label{sec:limitations}
\methodname{} enables safe, autonomous learning of multi-step mobile manipulation behaviors from third-person human video demonstrations. However, there are some limitations of the method that offer exciting avenues for future work. 
First, the simulated training data for the safety Q-functions must cover similar scenarios to those encountered during real-world exploration. Despite not requiring task-specific demonstrations or task successes in simulation, this does require some engineering of simulated tasks. This limitation could be alleviated by large-scale pretraining in simulation with a multitude of tasks and assets. 
Similarly, we study only a limited number of failure modes that we enumerate a priori. Scaling to other types of safety violations or task failures presents an opportunity for future work. 
Another limitation is that \methodname{} relies on initial correspondence in pose between the human and robot such that actions represented as relative motions yield similar trajectories. We obtain this correspondence by estimating the initial human pose in the map frame while the robot observes, and navigating the robot there at the start of the episode. 
\methodname{} also relies on grasp pose generation methods in order to obtain feasible grasps for the robot to explore ---  accurately inferring a suitable robot grasp from the demonstrated human grasp is an open challenge. 
A third limitation lies in the backtracking strategy: backtracking is able to reset some aspects of the environment, but is unable to recover from irreversible events like dropping objects. Integrating autonomous resetting frameworks~\cite{zhu2020ingredients,mendonca2024continuously, walke2023don} is a natural opportunity for future work. 
Finally, an exciting avenue for future work is improving the safety Q-functions given real-world data. While \methodname{}'s safety Q-functions demonstrate high accuracy and a low false negative rate, learning from occasional observed failures in the real-world could increase the robustness in future learning attempts. 

\section{Conclusion}
\label{sec:conclusion}

In this work, we present \methodname{}, a framework for safe, autonomous human-to-robot imitation for mobile manipulation tasks. 
\methodname{} effectively parses human demonstrations through a combination of pose tracking and VLM prompting, inferring both semantic segments and corresponding motions.
The framework then safely refines the demonstrated behavior by sampling and verifying actions using safety Q-functions pre-trained in simulation, backtracking and exploring new actions and grasp modes when necessary. 
The result is a policy that adapts the demonstrated behavior to the robot's morphology, and can be associated with the task to reduce exploration in future attempts. 
We validate \methodname{} on seven multi-step mobile manipulation tasks and find that it outperforms representative baselines in both task success and safety metrics, and is robust across different environments and human teachers. 
\arpit{Through our experiments we show that safety Q-functions learned in simulation outperform IL policies trained in simulation due to the higher data quality and quantity demands of the latter.} 
Taken together, our work is a promising step toward robots that can robustly learn new behaviors from human teachers in their own home environments. 

\section*{Acknowledgements} 
This work took place at the Robot Interactive Intelligence Lab (RobIn) at UT Austin. RobIn is supported in part by the College of Natural Sciences (CNS) Catalyst Grant (CAT-24-MartinMartin).
\bibliographystyle{unsrtnat}
\bibliography{references}

\clearpage
\newpage
\appendix

\subsection{Model Details}
\label{sec:app_details}

The architecture of the SQFs is a PointNet++~\cite{qi2017pointnet} encoder for processing point cloud input, 4-layer MLP encoder for the action and 4-layer MLP encoder for the proprioception (end-effector pose and the FT value). We categorize the continuous FT value into 6 bins for better sim-to-real transfer. Features from all the encoders are concatenated and input to an MLP head. The action space is cartesian delta positions of the end-effector with the delta actions in the range $[0.0, 0.15]$.

We define several possible task failures: arm collision, base collision, joint limit violation, force/torque threshold violation, grasp loss during articulated object manipulation, and object dropping.

\begin{figure}[t!]
    \centering
    \includegraphics[width=\columnwidth]{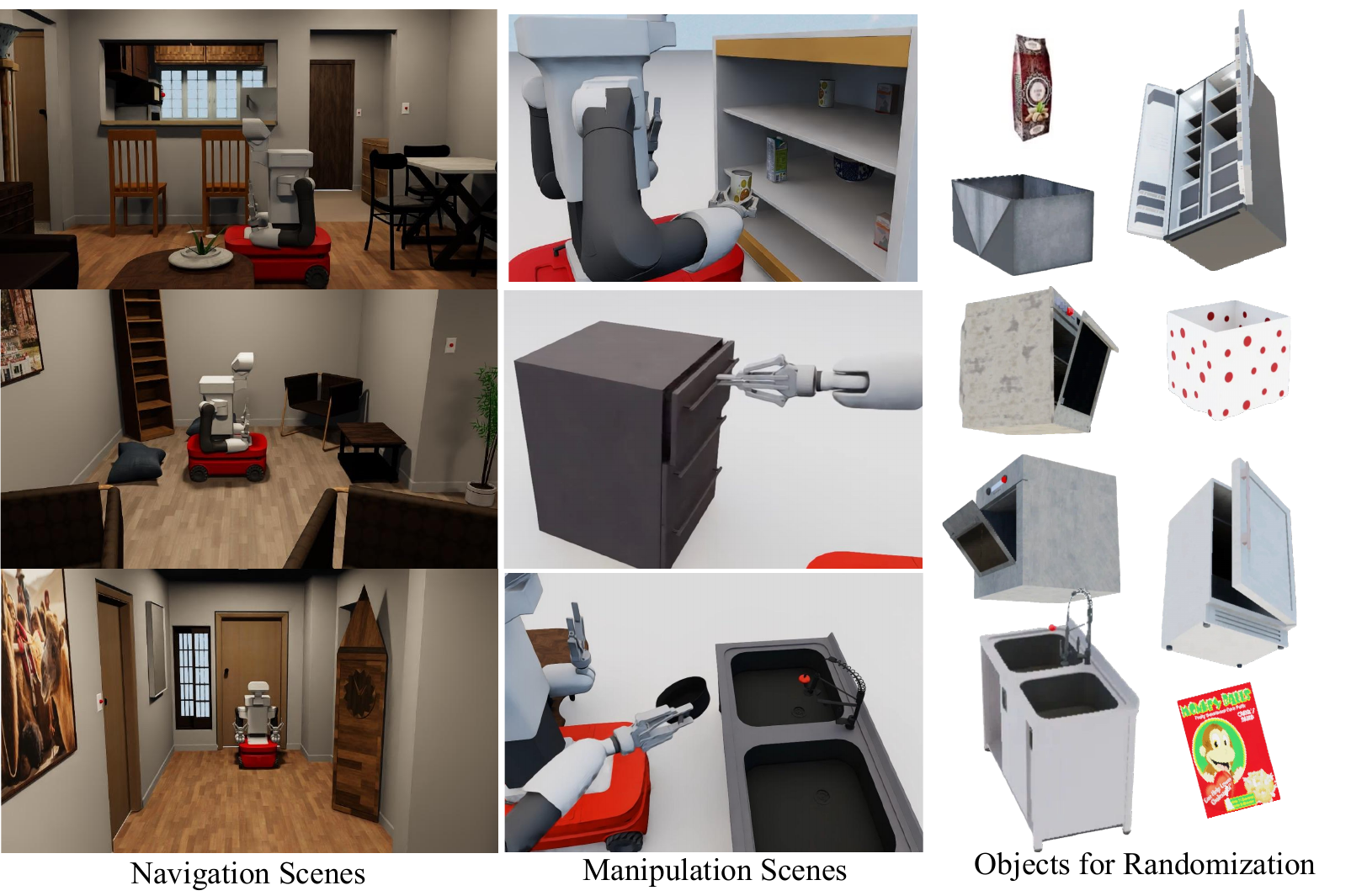}
    \vspace{-2em}
    \caption{Simulation training domains for the safety Q-function functions of \methodname{}. We generate random and targeted actions in multiple simulated scenes in OmniGibson~\cite{li2023behavior}. To increase generalization, the scenes are domain randomized using different objects and locations. We collect 43200 state-action pairs. Point clouds and robot proprioceptive signals are collected during data generation, as well as ground truth for different failure types, and used to train the ensemble of safety Q-function functions that \methodname{} uses for real-world exploration. The process is cheap and scalable and does not require solving the tasks in simulation.}
    \label{fig:sim}
\end{figure}

During data collection, trajectories are recorded along with the safety label for each state-action pair encountered (see Fig.~\ref{fig:sim}). The Q-functions are optimized using a cross-entropy loss. 
\begin{table}[h!]
\centering
\caption{Hyperparameters for safe exploration, SQF training, baselines, and video parsing.}
\begin{tabular}{ll}
\toprule
\textbf{Hyperparameter} & \textbf{Q-function} \\ 
\midrule
    SQF threshold $\epsilon$ & 0.7 \\
    Exploration Standard Deviation $\sigma$ & 0.05 \\
    Grasp Mode Clustering? & K-Mediods \\
    Batch Size & 16 \\ 
    Optimizer & Adam) \\
    Learning Rate & 1e-3 \\ 
    SQF Dataset Size & 9000 \\ 
    IL (all safe actions) Dataset Size & 23471 \\ 
    IL (successful episodes) Dataset Size & 4305 \\ 
    Nav/Stationary Pos Threshold 
     & 0.07 \\
     Nav/Stationary Ori Threshold 
     & 0.2 \\
\bottomrule
\end{tabular}
\vspace{-2em}
\end{table}



\begin{algorithm}
\caption{SafeMimic Execution Pseudocode}
\label{algo:algo}
\begin{algorithmic}[1]
\Require Segment videos $V_{1:N}$, 
video parser $f_\mathrm{parse}$, 
standard deviation $\sigma$, 
num action samples $m$,  
safety Q-functions $Q_\mathrm{safe}^{1:K}$, 
policy memory $\pi_\mathrm{mem}$
\For{segment $i=1...N$}
    \State $\tau_{human},$ segment\_type, $f_\mathrm{success} \gets \  f_{parse}(V_i)$
    \If{segment\_type is grasping}
        \State $G$ $\gets$ \texttt{sample\_grasp\_modes}()
        \If{$\pi_\mathrm{mem}$ exists for this segment} 
            \State $g_\mathrm{robot} \gets \pi_\mathrm{mem}(s_t)$
        \Else
            \State $g_\mathrm{human} \gets \tau_\mathrm{human}\left[-1\right]$
            \State $g_\mathrm{robot} \gets$ \texttt{closest}($G, g_\mathrm{human}$)
        \EndIf
        \State \texttt{execute}($g_\mathrm{robot}$) \State $G$.\texttt{pop}($g_\mathrm{robot}$)
    \Else \Comment{Segment is nav or manipulation}
    \If{$\pi_\mathrm{mem}$ exists for this segment} 
            \State Sample $\tau_{1:m} \sim \pi_\mathrm{mem}(s_t)$
        \Else
        \State Sample $\tau_{1:m} \sim \mathcal{N}(\tau_\mathrm{human}, \sigma)$ 
        \EndIf
        \State segment\_success $\gets$ False
        \While{True}
            \State segment\_success $\gets$ \texttt{safe\_explore}($\tau_{1:m}$)
            \If{segment\_success}
                \State \textbf{break}\Comment{Go to next segment}
            \EndIf
            \If{$G$ is not empty} 
                \State $g_{robot} \sim G $ \Comment{Explore new grasp}
                \State \texttt{execute}($g_{robot}$)
                \State $G$.\texttt{pop}($g_{robot}$)
            \EndIf
        \EndWhile
    \EndIf
\If{not segment\_success} \Return False
\EndIf
\EndFor
\State \Return segment\_success
\Procedure{safe\_explore}{$\tau_{1:m}$}
\If{len($\tau_{1:m}$) $== 0$} \Comment{End of trajs}
\State \Return $f_{\mathrm{success}}()$
\EndIf

\State Filter $\tau_{1:m}$ if first action $Q_\mathrm{safe}^{k}(s_t, a_0) < \epsilon \ \forall \ k$ 
\State Sort $\tau_{1:m}$ by first action $\max_kQ_\mathrm{safe}^{1:K}(s_t, a_0)$ ascending
\For{each $a_0$ in $\tau_{1:m}$}
    \State \texttt{execute}($a_0$)
    \State success $\gets$ \texttt{safe\_explore}(\texttt{next}($\tau_{1:m}$))
    \If{success} \Return True
    \EndIf
    \State \texttt{execute}($-a_0$) \Comment{Backtrack one step}
\EndFor

\State \Return False

\EndProcedure

\end{algorithmic}
\end{algorithm}

%

\subsection{Human Video Parsing}
\label{sec:app_segmentation}
\begin{figure}[t!]
    \centering
    \includegraphics[width=\columnwidth,trim={0 2cm 0 2cm},clip]{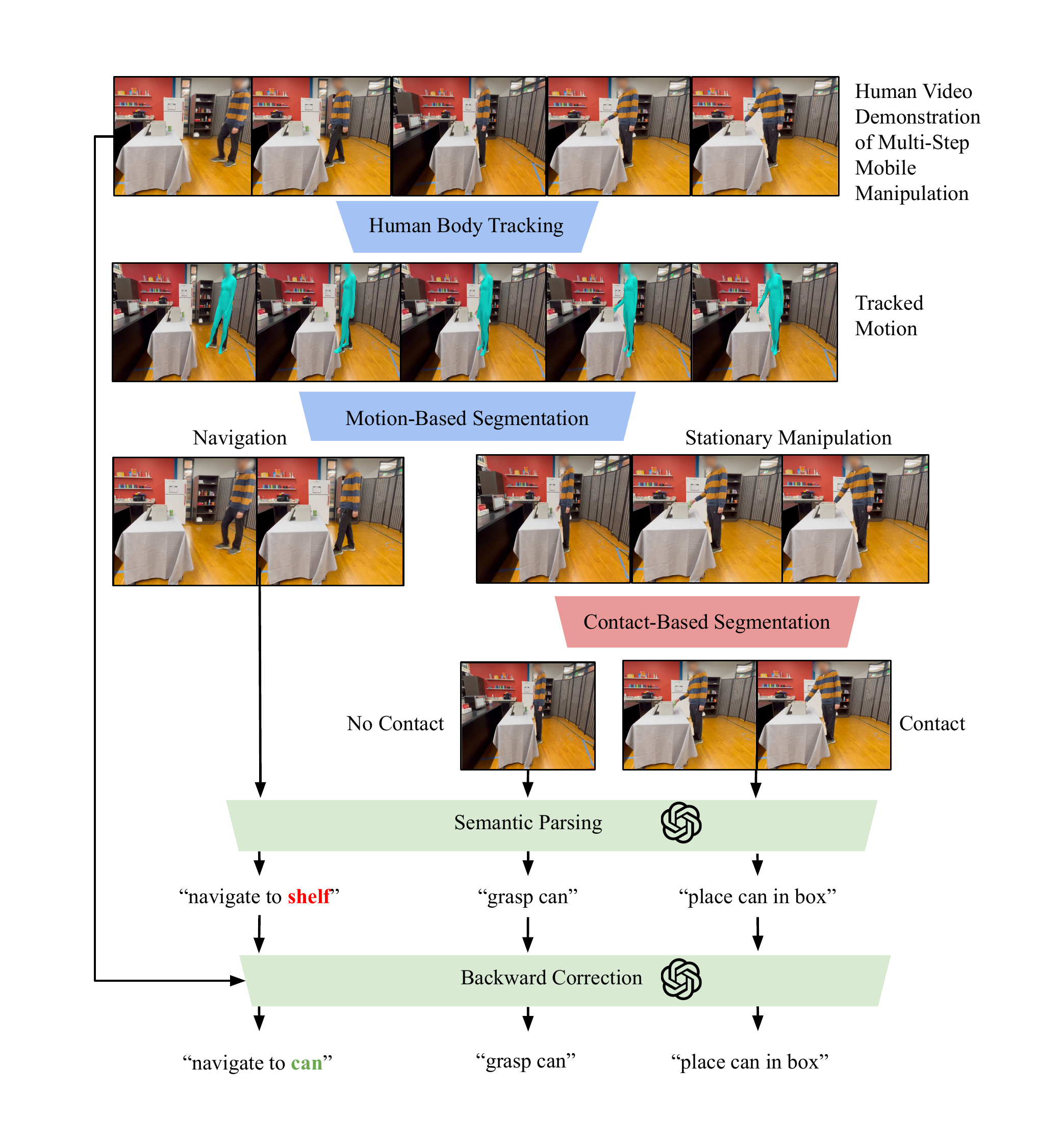}
    \caption{Human video parsing by \methodname. An initial RGB-D video demonstration is processed by \methodname{} using a body tracking solution to obtain segments where the human is either navigating or performing stationary manipulation. 
    Initial stationary manipulation segments are further segmented based on changes in the hand-object contact state.
    For all segments, the navigation and the manipulation ones, \methodname{} prompts a VLM that provide a semantic description of the actions. 
    \methodname{} refines the semantic labeling results by reprompting the VLM for backward consistency between labels.
    The segmentation (motion and semantic labels) are used by \methodname{} to refine and adapt the actions in a safe and autonomous manner.}
    \label{fig:app_segmentation}
\end{figure}

\methodname's{} video parsing module extracts semantic task segments, including the semantic goal of each segment and the human actions to be adapted to the robot. The process is depicted in Fig.~\ref{fig:app_segmentation}. First, \methodname{} extracts human motion from the RGB-D video using the combination of body~\citet{pavlakos2024reconstructing} and hand~\citet{ye2023slahmr} tracking. 
Then, \methodname{} performs a coarse segmentation into navigation and stationary manipulation segments by thresholding the human hip motion. 

The stationary segments are refined into segments with hand contact and without hand contact using a coarse-to-fine second segmentation. 
First, \methodname{} queries a VLM at a low frame rate ($3$ fps) to determine whether the human hand is in contact with an object. 
We use OpenAI's GPT-4o~\cite{achiam2023gpt} for segmentation and later labeling. 
To obtain a more precise initial contact frame estimation, we refine the segmentation using a visual contact detector~\citet{Shan20} around the frames where the VLM detects a change in contact state. 

For all segments obtained (navigation and stationary manipulation, with and without contact), \methodname{} identifies and labels the semantic goal that will be used to enable autonomous success detection. 
Labeling the segments is performed with VLM queries that classify the segments into a set of possible actions and the objects the actions apply to, e.g., \textit{navigate to a can}.
Table~\ref{possible_actions} includes the list of actions the VLM classifies the segments on.

\begin{table}[ht]
\caption{Segment Types and Possible Actions}
\label{possible_actions}
\centering
\begin{tabular}{|l|l|}
\hline
\textbf{Segment Type} & \textbf{Possible Actions} \\
\hline
Navigation & 
\begin{tabular}[c]{@{}l@{}} Navigating to [object]\\ Navigating with [object$_1$] to [object$_2$]\\ \end{tabular} \\
\hline
Grasping & 
\begin{tabular}[c]{@{}l@{}} Reach for and grasp [object]\\ \end{tabular} \\
\hline
Manipulation & 
\begin{tabular}[c]{@{}l@{}} Pick [object]\\ 
Place [object$_1$] in [object$_2$] \\ Open [object]\\ Close [object]\\ Wipe [object$_1$] with [object$_2$] \end{tabular} \\
\hline
\end{tabular}
\end{table}

Since the segments are labeled independently, we observe some temporal inconsistencies in the results. 
We removed this effect through a backward correction method for the semantic descriptions by prompting again the VLM with all the descriptions in order, as well as frames from the entire video, requesting it to correct any inconsistencies between consecutive actions and objects. 
This helps significantly in situations where objects are inconsistent across descriptions, e.g., in the task of \texttt{store\_in\_drawer}, semantic segments include \textit{navigating to drawer} and \textit{opening drawer}. However, the initial labeling may result in a first semantic label \textit{``navigating to sink''} due to a sink being next to the drawer. 
By having access to the entire video and future labels, the VLM detects and corrects the wrong initial description. 

The segments obtained in this process, including both the human motion and the semantic subgoal label, are transformed into the robot's initial motion for \methodname{} to refine and adapt through safe and autonomous exploration.

\begin{figure}[t!]
    \centering
    \centering
    \includegraphics[width=0.32\columnwidth]{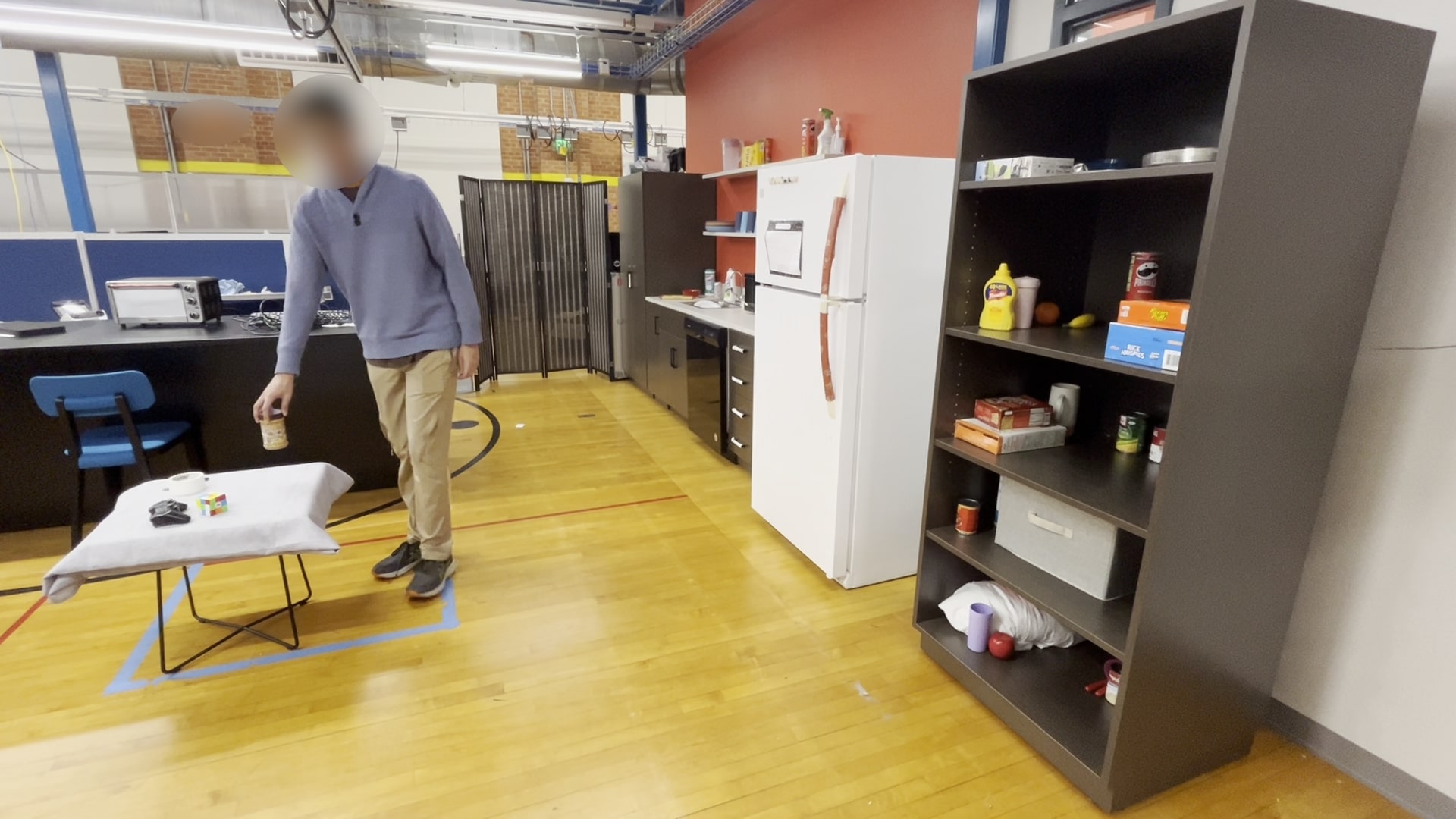}%
    \hfill%
    \includegraphics[width=0.32\columnwidth]{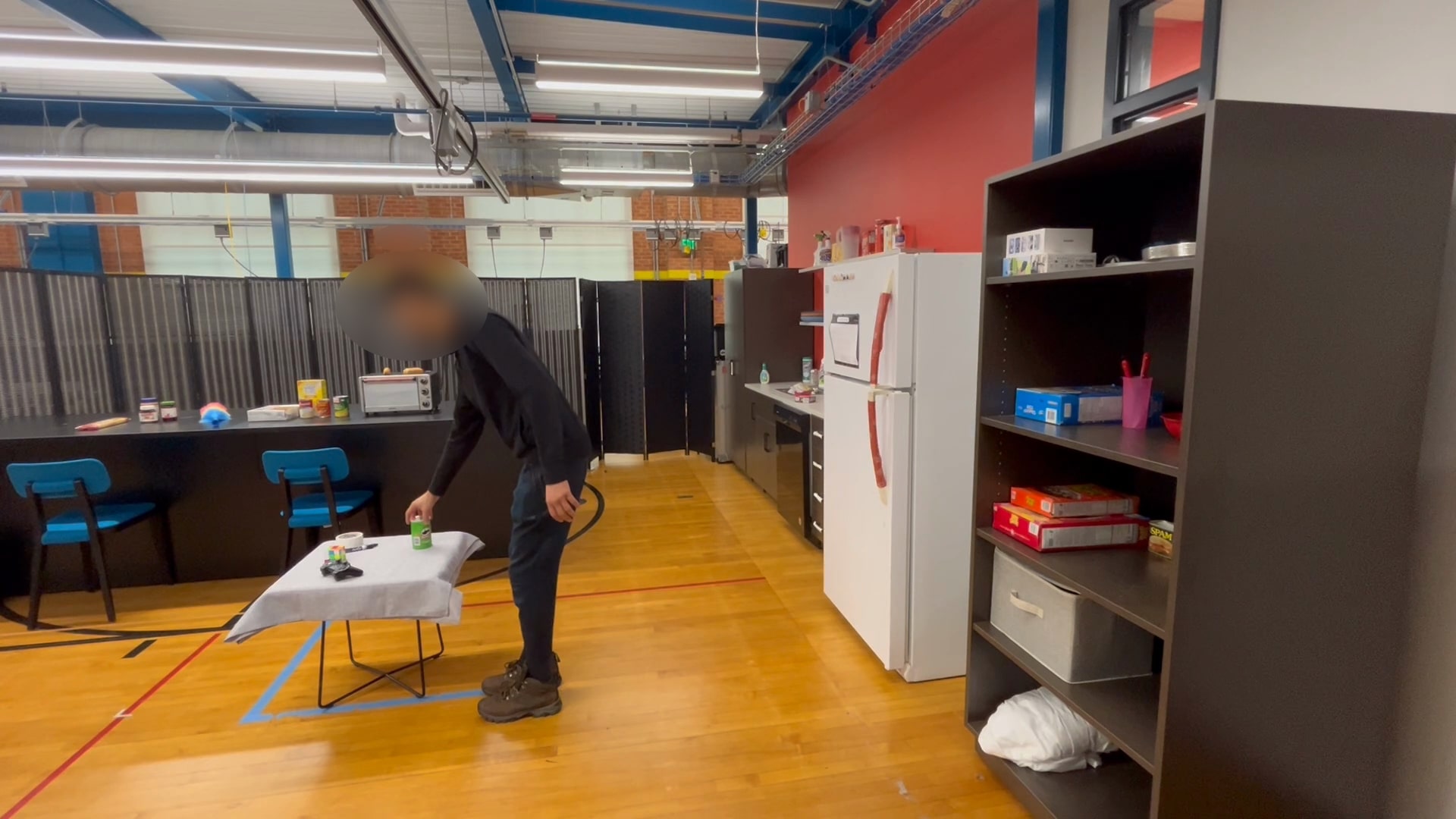}%
    \hfill%
    \includegraphics[width=0.32\columnwidth]{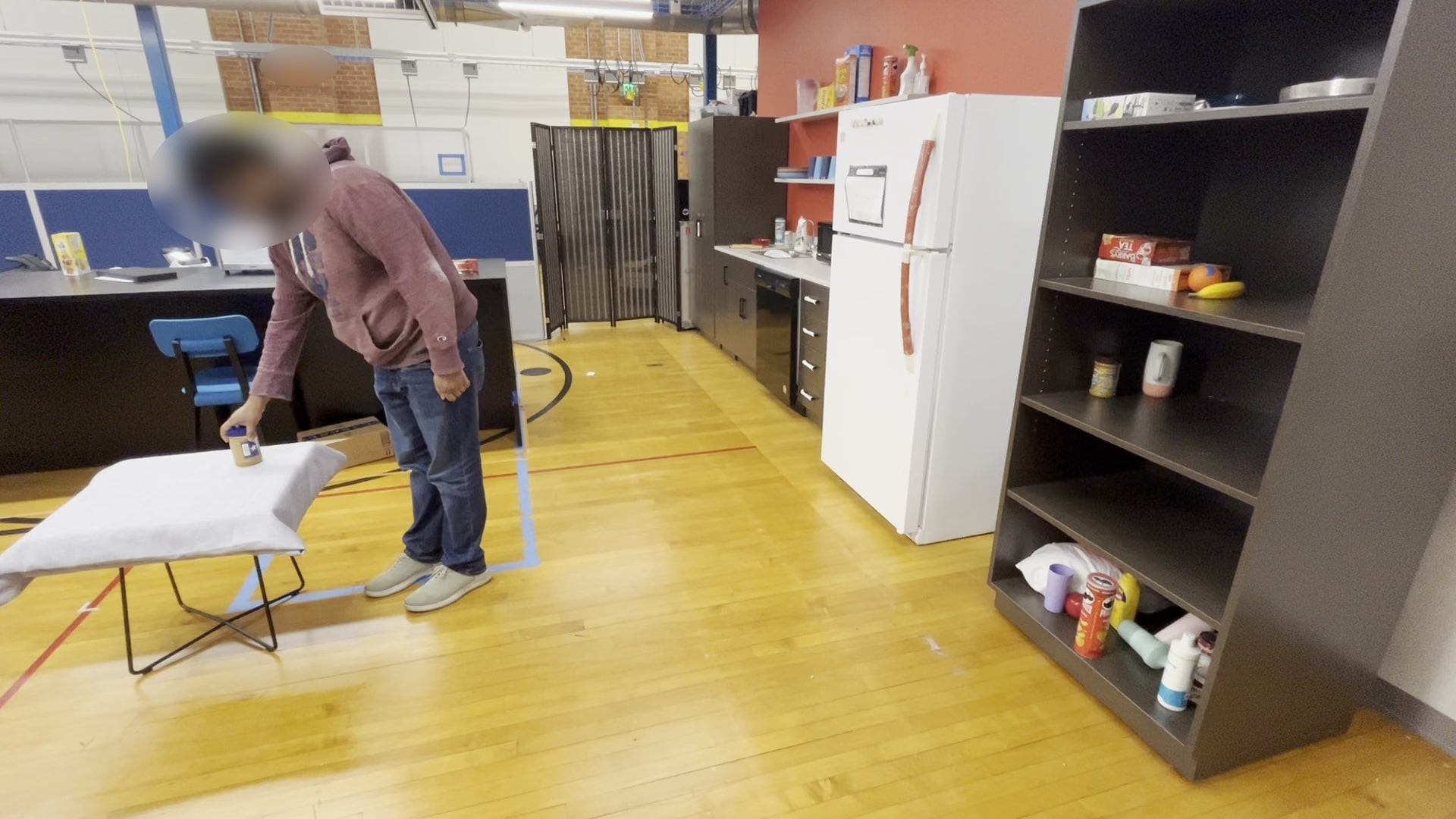}\\\vspace{5pt}
    \includegraphics[width=0.32\columnwidth]{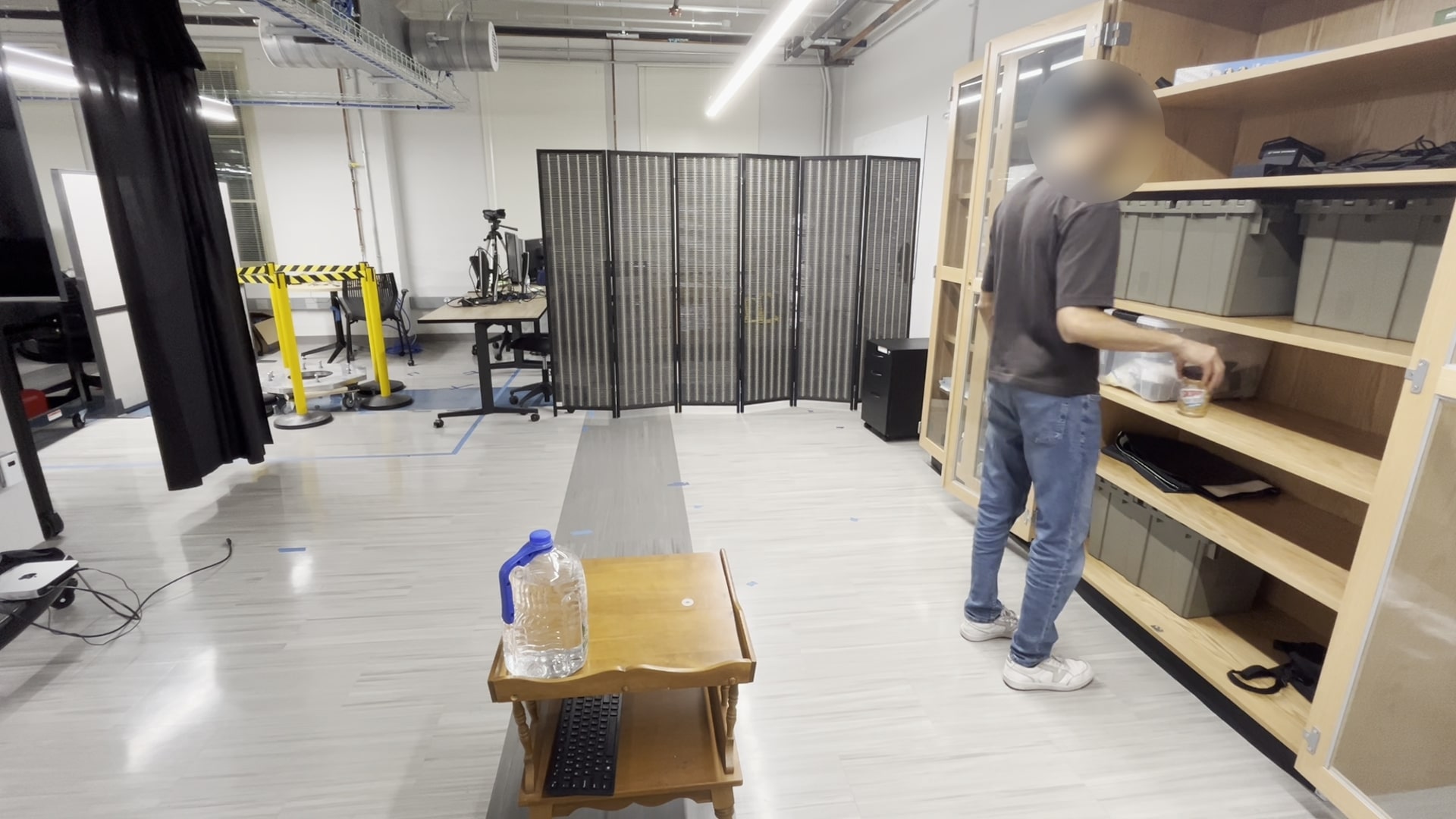}%
    \hfill%
    \includegraphics[width=0.32\columnwidth]{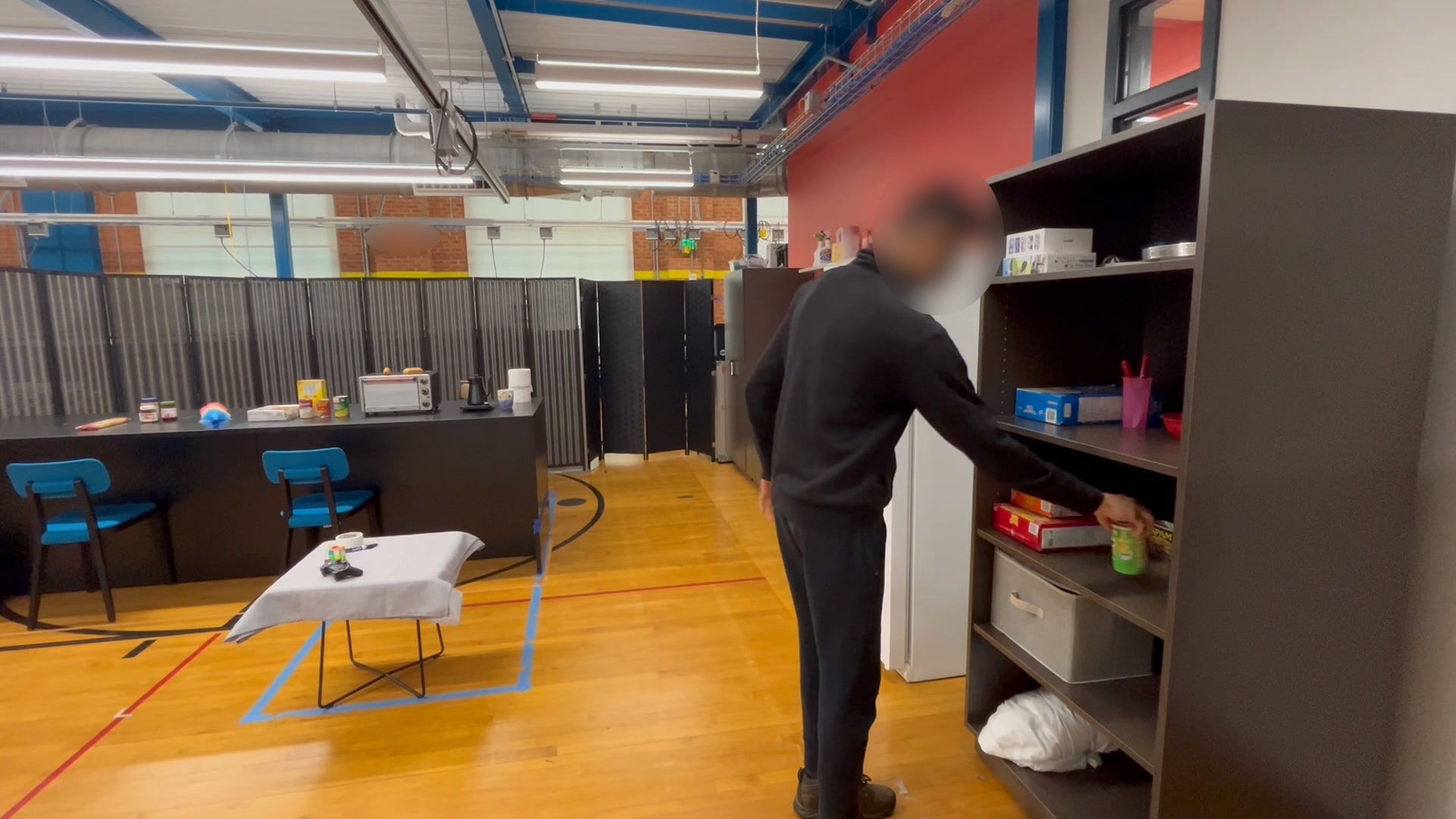}%
    \hfill%
    \includegraphics[width=0.32\columnwidth]{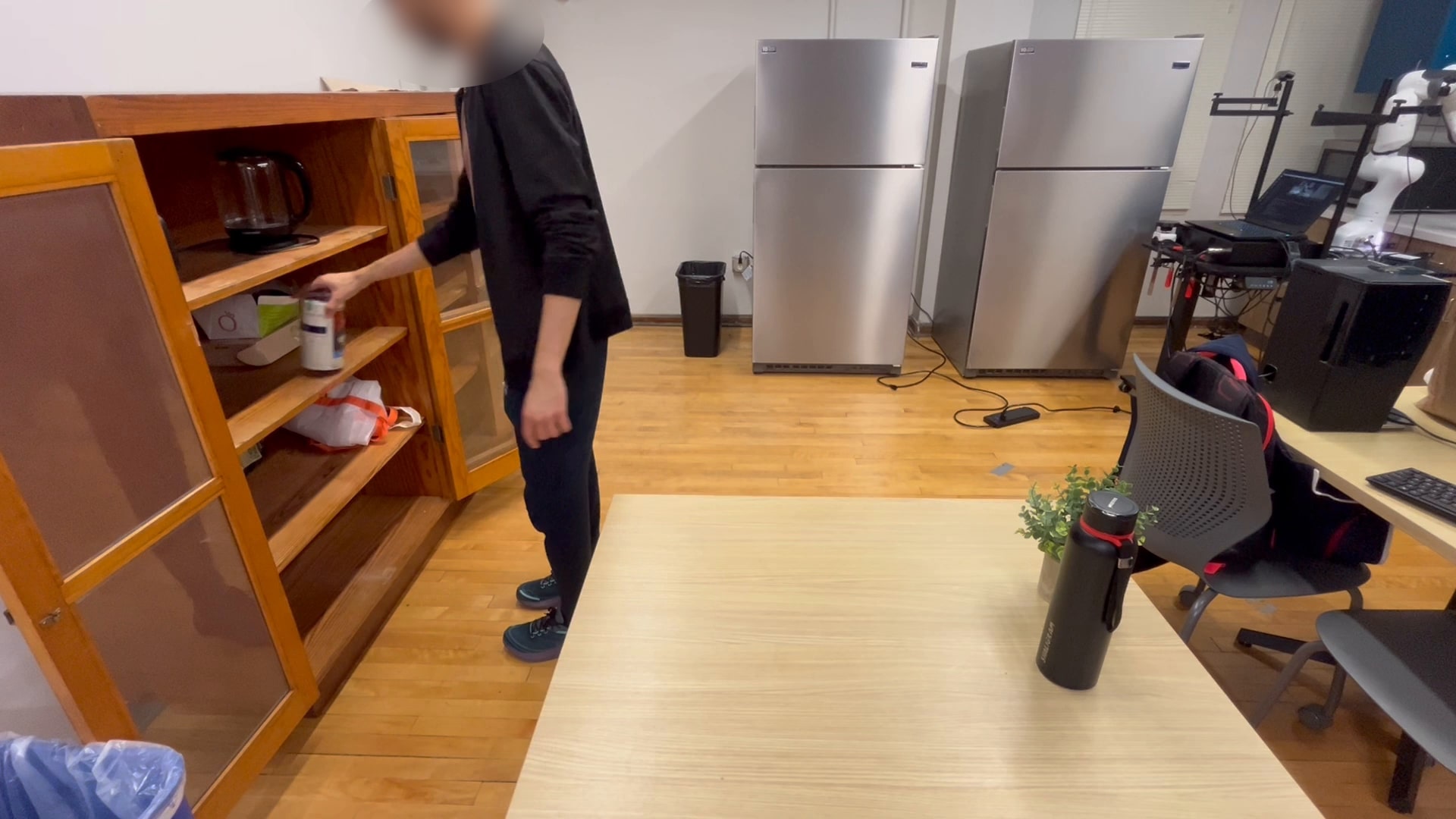}
    \caption{Snapshots of three different human demonstrations of the same task (top), and demonstrations of the task in three different environments (bottom). We evaluate \methodname{}'s abilities to parse initial demonstrations of different humans, and in different environments. 
    The performance of \methodname{} is not affected (Sec.~\ref{sec:results}), indicating that our method is robust to variations in these dimensions and is able to explore and adapt actions in different conditions.}
    \label{fig:humansenvs}
\end{figure}

\begin{figure}[t!]
    \centering
    \includegraphics[width=\columnwidth]{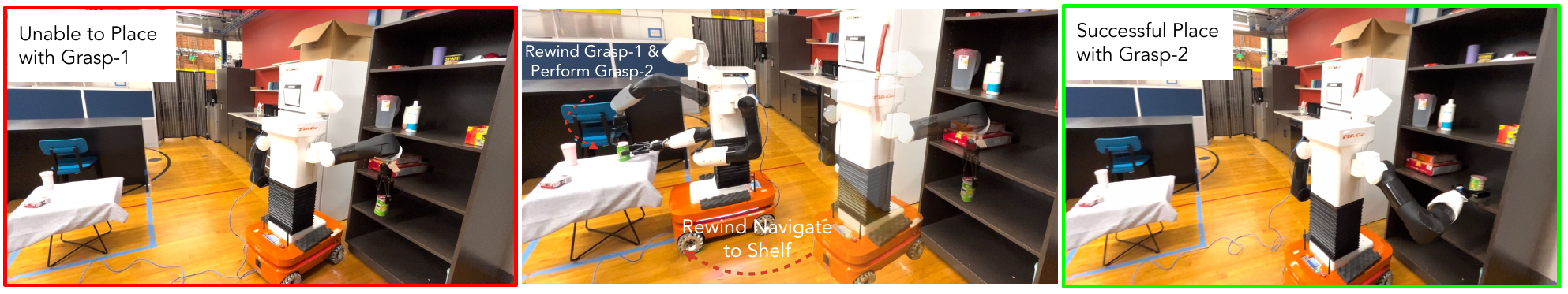}
    \caption{Backtracking actions in \methodname{}. Example sequence of backtracking behavior in \texttt{Shelving} task. When no safe samples are available, or when a segment exploration ends without success, \methodname{} backtracks to previous states by undoing the last actions, possibly stepping back to explore a different grasping mode. This simple but effective mechanism enables autonomous exploration and adaptation of human motion for multi-step mobile manipulation tasks with \methodname{}.}
    \label{fig:backtracking}
\end{figure}

\begin{figure}[t!]
    \centering
    \centering
    \includegraphics[width=0.32\columnwidth]{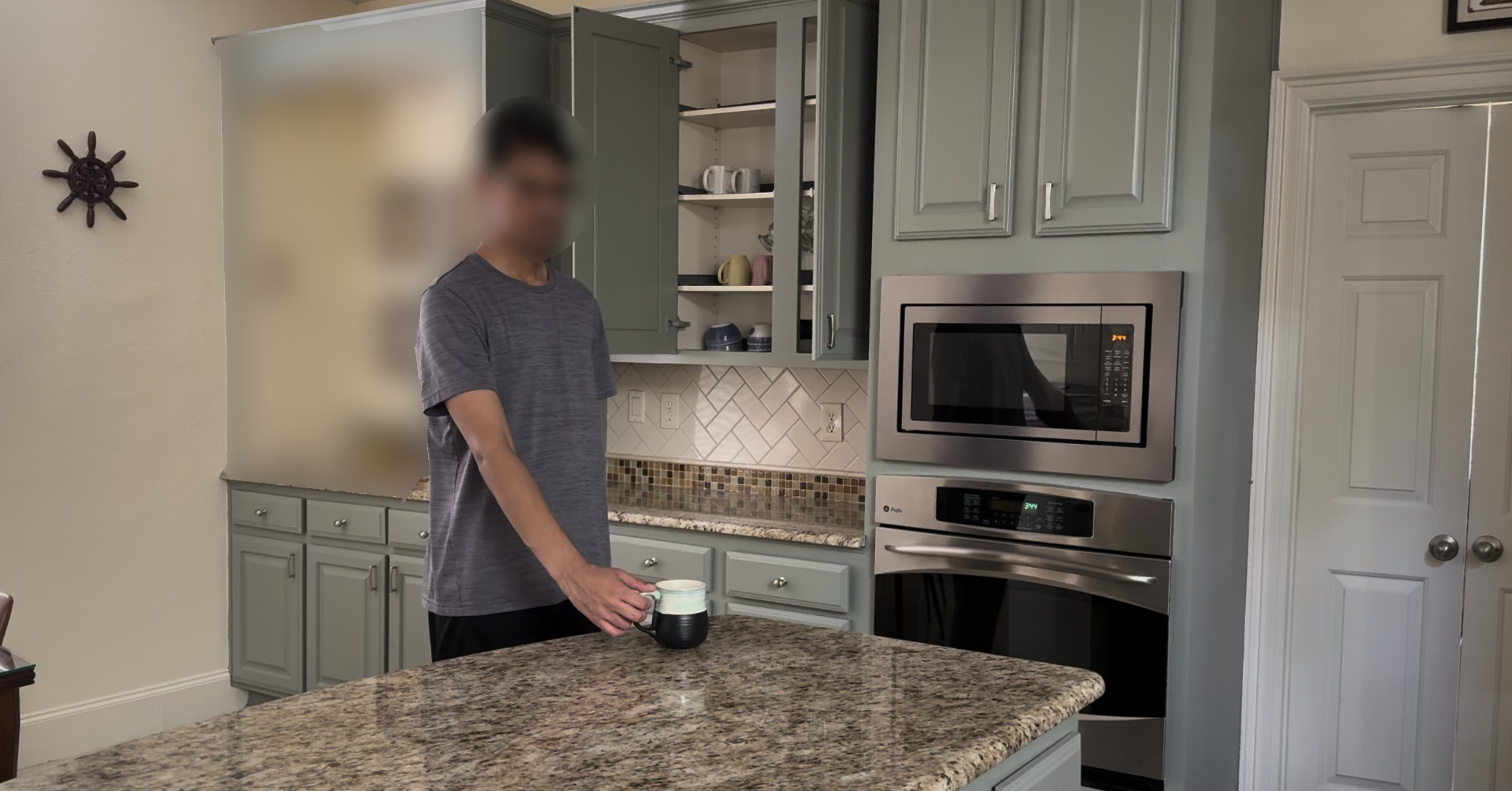}%
    \hfill%
    \includegraphics[width=0.32\columnwidth]{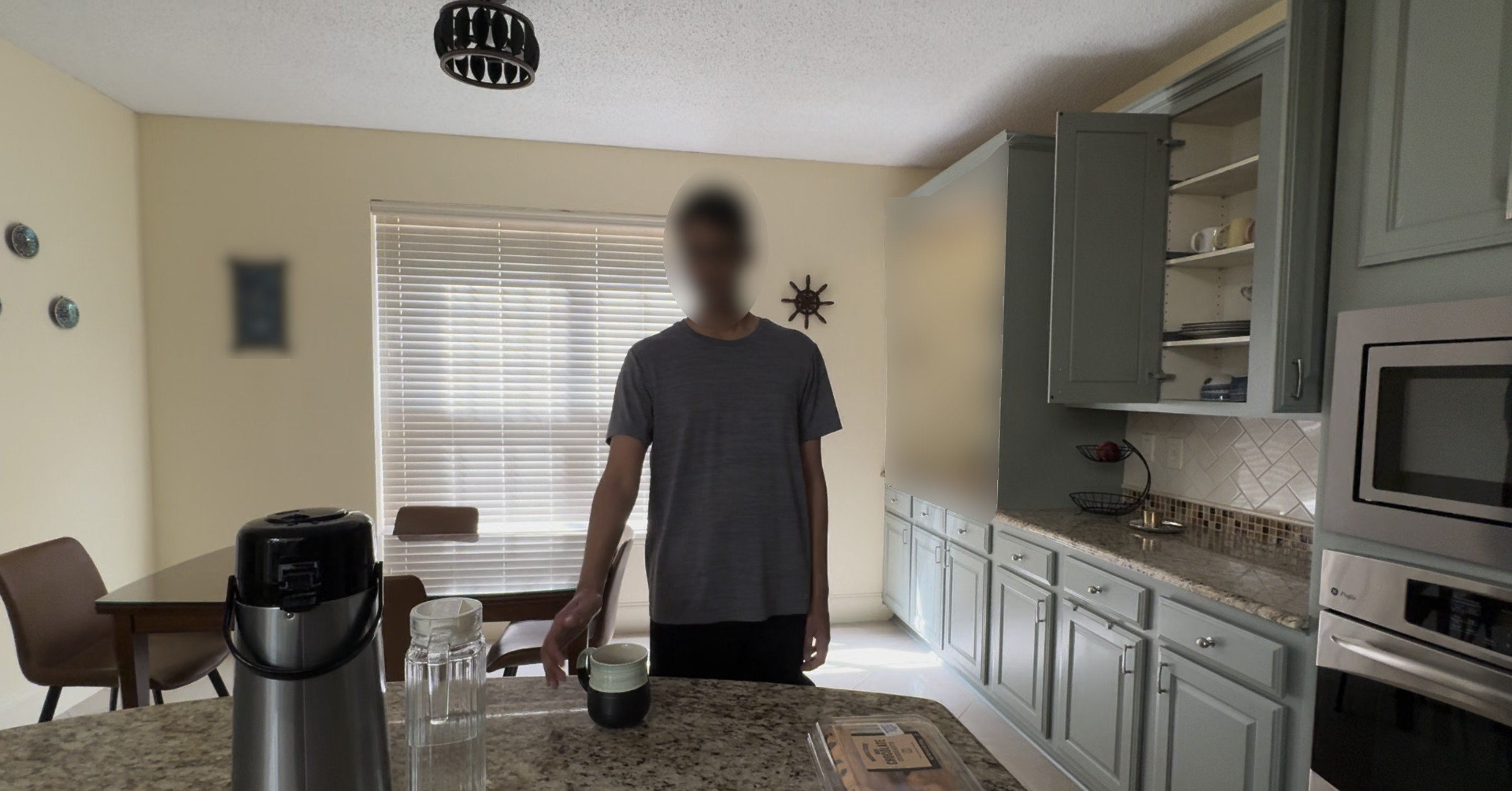}%
    \hfill%
    \includegraphics[width=0.32\columnwidth]{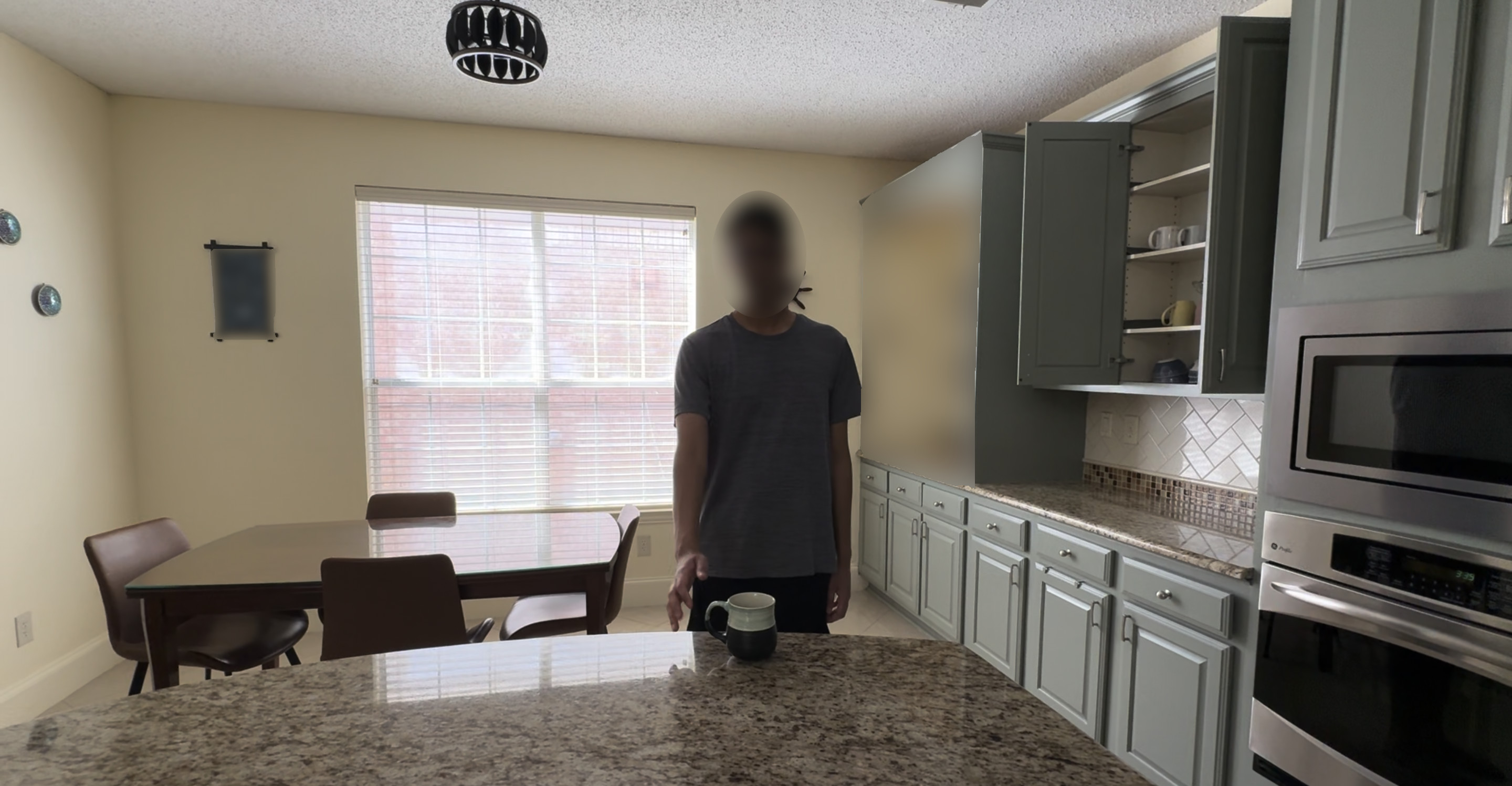}\\\vspace{5pt}
    \includegraphics[width=0.32\columnwidth]{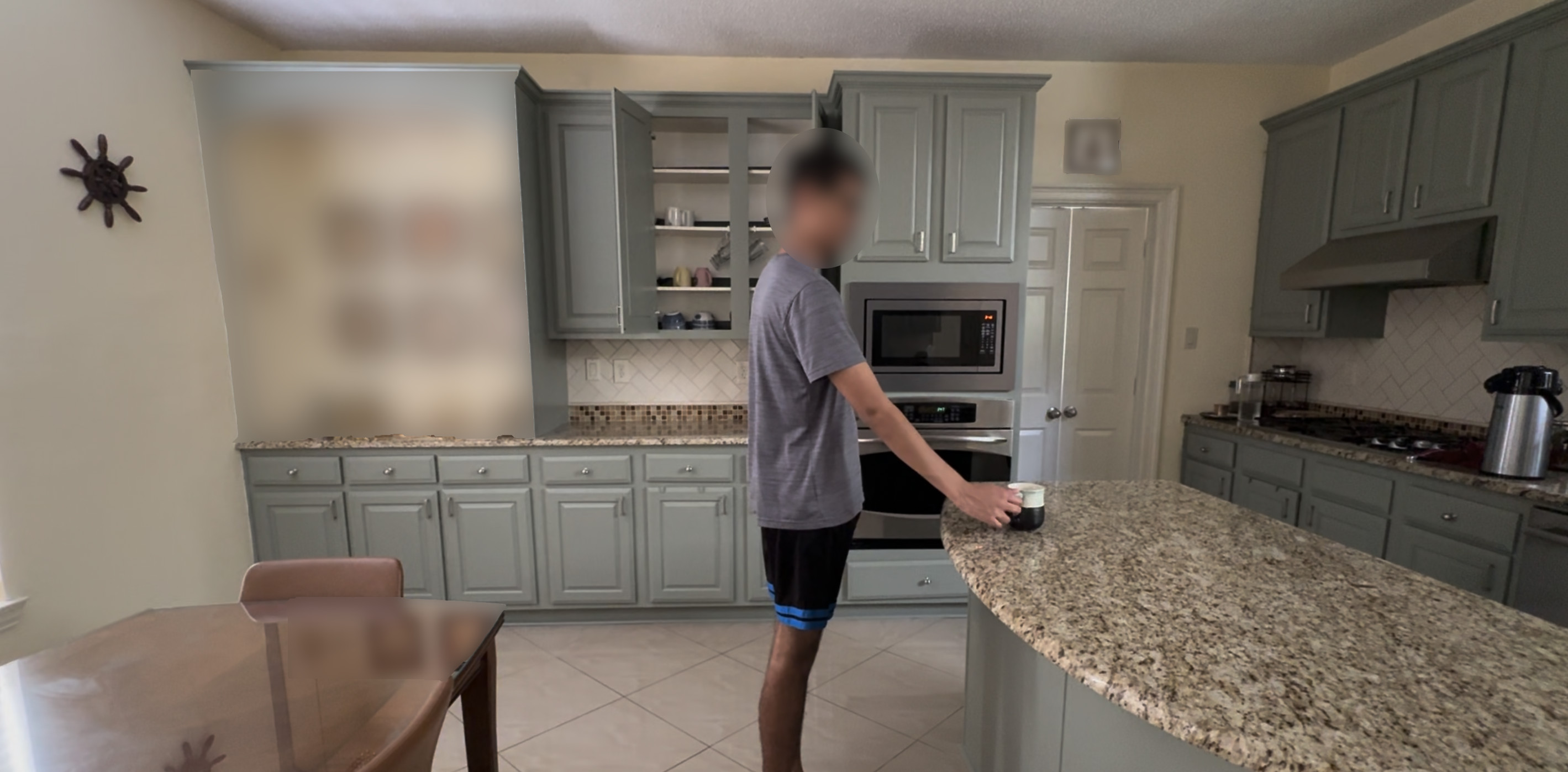}%
    \hfill%
    \includegraphics[width=0.32\columnwidth]{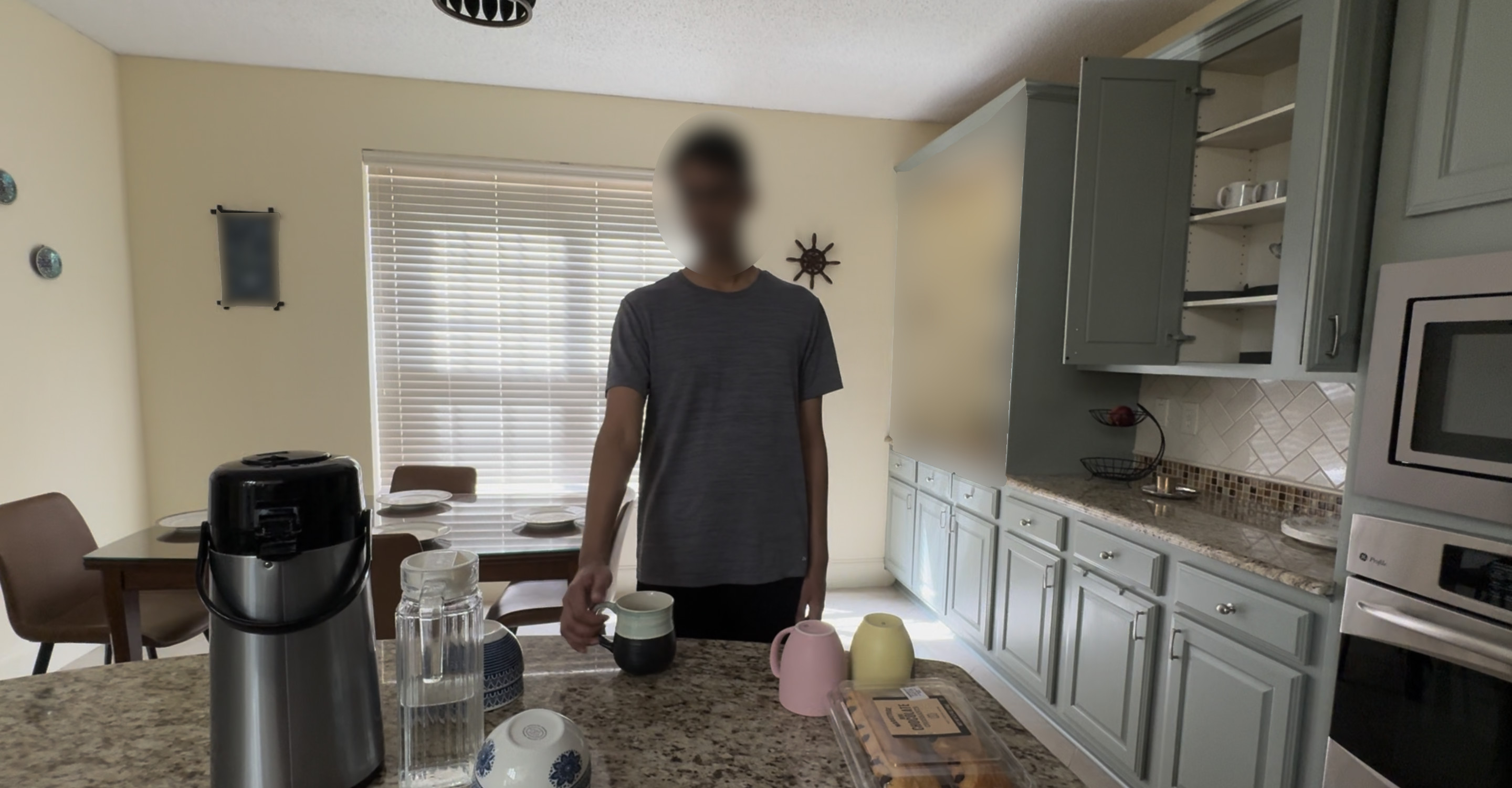}%
    \hfill%
    \includegraphics[width=0.32\columnwidth]{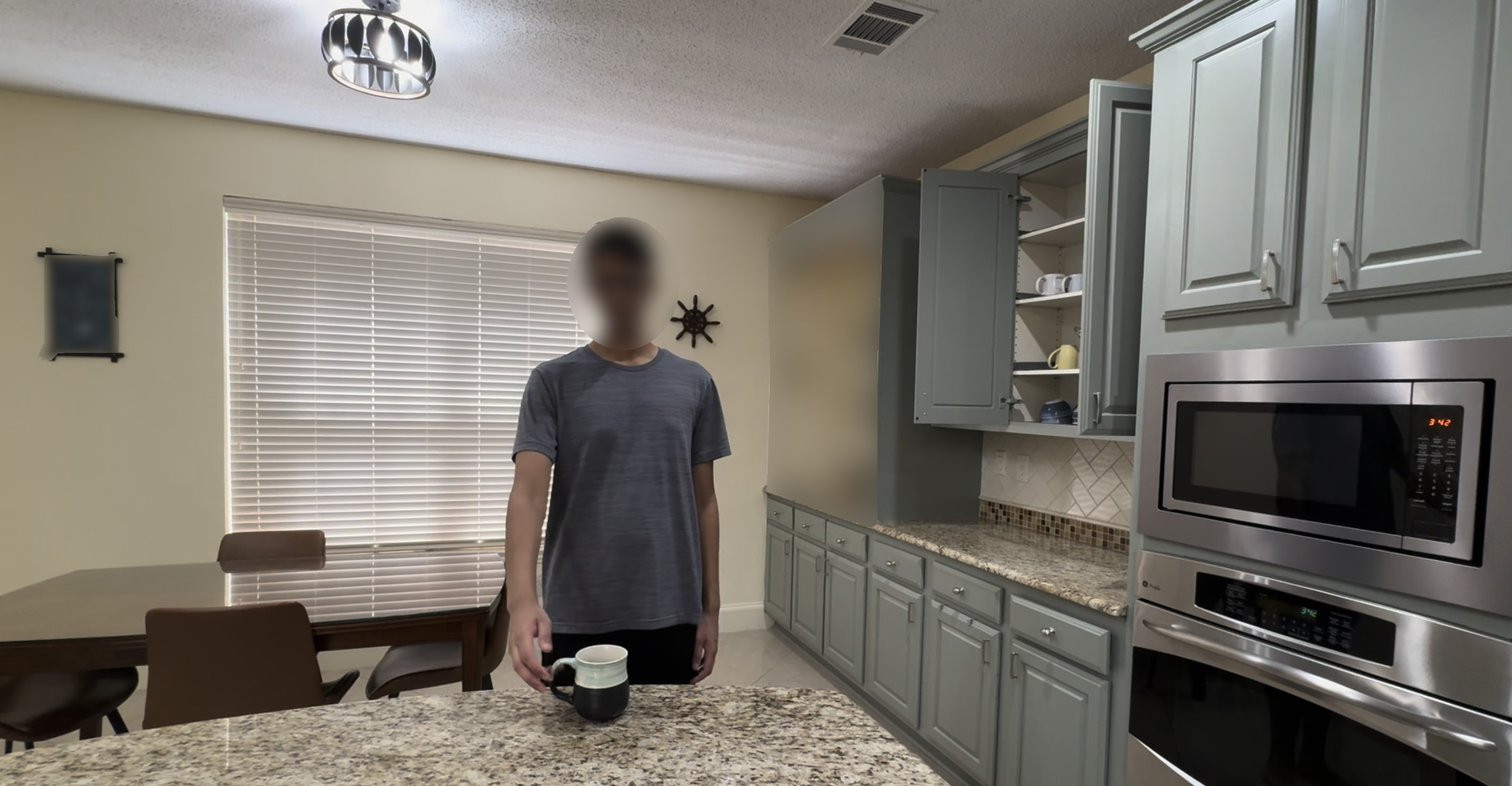}
    \caption{Snapshots of the videos we evaluate \methodname{}'s video parsing module on in \ref{sec:video_seg_results}. We test the module on videos with different camera angles (left column), different levels of clutter (middle column), and different lighting conditions (right column). Our results indicate that \methodname{} is robust to videos in varying and challenging environments.}
    \label{fig:video_seg_test_fig}
\end{figure}

\subsection{Robustness of the Human Video Parsing Module}
\label{sec:video_seg_results}
We have already shown in Fig. \ref{fig:humansenvs} that \methodname{}'s video parsing module is robust to videos taken in different environments with different human demonstrators. However, to further demonstrate the robustness of the module, we conducted the following tests in challenging environments:
We recorded videos of humans performing the \textit{shelving} task from different camera angles, different lighting conditions and with ``few'' (1-3), ``several'' (3-5), and ``many'' ($>$5) distractor objects, and measured success rate as the percent of correctly identified segments in the video. We observe \textbf{86\%} success across different camera angles, \textbf{100\%} across different lighting conditions and \textbf{88\%} across different levels of clutter, indicating that our activity segmentation is a robust first module for \methodname{}.

\end{document}